\documentclass[twocolumn, 10pt]{IEEEtran}
\usepackage{url}
\usepackage{graphicx}
\usepackage{subcaption}
\usepackage{amssymb}
\usepackage{amsmath}
\usepackage{mathtools}
\usepackage{dsfont}
\usepackage{cite}
\usepackage{stfloats}
\usepackage{psfrag}
\usepackage[mathscr]{euscript}
\usepackage{acronym}  
\usepackage{booktabs}
\usepackage{bbm}
\usepackage{algorithm}
\usepackage{algcompatible}
\usepackage{algorithmicx}
\usepackage{multirow}

\usepackage{float}
\usepackage{balance}

\acrodef{CCDF}{complementary cumulative distribution function}
\acrodef{CF}{characteristic function}
\acrodef{PPP}{Poisson point processe}
\acrodef{RV}{random variable}
\acrodef{i.i.d.}{independent and identically distributed}
\acrodef{PDF}{probability distribution function}
\acrodef{CDF}{cumulative distribution function}
\acrodef{ch.f.}{characteristic function}
\acrodef{AWGN}{additive white Gaussian noise}
\acrodef{SNR}{signal-to-noise ratio}
\acrodef{LRT}{likelihood ratio test}
\acrodef{DRT}{distance ratio test}
\acrodef{GLRT}{generalized likelihood ratio test}
\acrodef{CRLB}{Cram\'{e}r-Rao lower bound}
\acrodef{CRB}{Cram\'{e}r-Rao bound}
\acrodef{ZZLB}{Ziv-Zakai lower bound}
\acrodef{ZZB}{Ziv-Zakai bound}
\acrodef{LOS}{line-of-sight}
\acrodef{ToF}{time-of-flight}
\acrodef{NLOS}{non-line-of-sight}
\acrodef{GDOP}{geometric dilution of precision}
\acrodef{GPS}{Global Positioning System}
\acrodef{FIM}{Fisher information matrix}
\acrodef{PEB}{position error bound}
\acrodef{SPEB}{squared position error bound}
\acrodef{TOA}{time-of-arrival}
\acrodef{TOF}{time-of-flight}
\acrodef{WSN}{wireless sensor network}
\acrodef{MAC}{medium access control}
\acrodef{RSS}{received signal strength}
\acrodef{WAF}{wall attenuation factor}
\acrodef{TDOA}{time difference-of-arrival}
\acrodef{RF}{radiofrequency}
\acrodef{RTT}{round-trip time}
\acrodef{AOA}{angle-of-arrival}
\acrodef{MF}{matched filter}
\acrodef{ED}{energy detector}
\acrodef{ML}{maximum likelihood}
\acrodef{MSE}{mean-square error}
\acrodef{RMSE}{root-mean-square error}
\acrodef{LEO}{localization error outage}
\acrodef{ppm}{part-per-million}
\acrodef{ACK}{acknowledge}
\acrodef{UWB}{Ultrawide bandwidth}
\acrodef{TNR}{threshold-to-noise ratio}
\acrodef{LS}{least squares}
\acrodef{IR-UWB}{impulse radio UWB}
\acrodef{FCC}{Federal Communications Commission}
\acrodef{TH}{time-hopping}
\acrodef{PPM}{pulse position modulation}
\acrodef{MUI}{multi-user interference}
\acrodef{PDP}{power delay profile}
\acrodef{BPZF}{band-pass zonal filter}
\acrodef{SIR}{signal-to-interference ratio}
\acrodef{SINR}{signal-to-interference-plus-noise ratio}
\acrodef{RFID}{radio frequency identification}
\acrodef{WPAN}{wireless personal area network}
\acrodef{WWB}{Weiss-Weinstein bound}
\acrodef{DP}{direct path}
\acrodef{MF}{matched filter}
\acrodef{MMSE}{minimum-mean-square-error}
\acrodef{SBS}{serial backward search}
\acrodef{SBSMC}{serial backward search for multiple clusters}
\acrodef{NBI}{narrowband interference}
\acrodef{WBI}{wideband interference}
\acrodef{INR}{interference-to-noise ratio}
\acrodef{CR}{channel response}
\acrodef{CIR}{channel impulse response}
\acrodef{CR}{channel  response}
\acrodef{RADAR}{radar}
\acrodef{MUR}{Multistatic radar}
\acrodef{JBSF}{jump back and search forward}
\acrodef{HDSA}{high-definition situation-aware}
\acrodef{RRC}{root raised cosine}
\acrodef{ST}{simple thresholding}
\acrodef{BTB}{Bellini-Tartara bound}
\acrodef{P-Max}{$P$-Max}  
\acrodef{MIMO}{multiple-input multiple-output}
\acrodef{MAP}{maximum a posteriori}
\acrodef{FG}{factor graph}
\acrodef{OP}{outage probability}
\acrodef{WED}{wall extra delay}
\acrodef{RMS}{root mean square}
\acrodef{SPAWN}{sum-product algorithm over a wireless network}
\acrodef{MDD}{minimum distance distribution}
\acrodef{MAP}{maximum a posteriori probability}
\acrodef{SAP}{small cell access point}
\acrodef{UE}{user equipment}
\acrodef{MBS}{macro cell base station}
\acrodef{UER}{\ac{UE} Relay}
\acrodef{D2D}{device-to-device}
\acrodef{MBS}{macro base station}
\acrodef{CSI}{channel state information}
\acrodef{OGR}{outage guard region}
\acrodef{FUR}{feasible UER region}
\acrodef{EHR}{energy harvesting region}
\acrodef{EH}{energy harvesting}
\acrodef{D2D-EHSN}{D2D communication provided \ac{EH} small cell network}
\acrodef{D2D-EHHN}{D2D communication provided \ac{EH} heterogeneous network}
\acrodef{3GPP}{3rd Generation Partnership Project}
\acrodef{BS}{base station}
\acrodef{DF}{decode and forward}
\acrodef{CCDF}{complementary cumulative distribution function}
\acrodef{ZF}{zero forcing}
\acrodef{RZF}{regularized zero forcing}
\acrodef{WLLN}{weak law of large number}
\acrodef{SLLN}{strong law of large numbers}
\acrodef{TDD}{Time-division duplex}
\acrodef{EE}{energy efficiency} 
\acrodef{HetNet}{heterogeneous network} 
\acrodef{SCP}{Single Cell Processing}
\acrodef{CBF}{Coordinated Beamforming}
\usepackage{color}
\usepackage{dsfont}
\usepackage{bbm}

\DeclareMathAlphabet{\mathsf}{OML}{cmbr}{m}{it}

\newtheorem{theorem}{\bf Theorem}
\newtheorem{lemma}{\bf Lemma}

\newtheorem{remark}{\bf Remark}

\newtheorem{assumption}{\bf Assumption}

\newcommand{\bd}{\begin{description}}
\newcommand{\ed}{\end{description}}
\newcommand{\be}{\begin{enumerate}}
\newcommand{\ee}{\end{enumerate}}
\newcommand{\bi}{\begin{itemize}}
\newcommand{\ei}{\end{itemize}}
\newcommand{\bl}{\begin{list}}
\newcommand{\el}{\end{list}}
\newcommand{\bt}{\begin{tabbing}}
\newcommand{\et}{\end{tabbing}}

\newcommand{\paperTitle}{Timely Parameter Updating in Over-the-Air Federated Learning }

\begin{document}

{
\title{\paperTitle}

\author{

	\vspace{0.2cm}
	  Jiaqi~Zhu,     
    Zhongyuan Zhao, \textit{Senior Member, IEEE}, 
    Xiao Li, \textit{Member, IEEE}, 
    Ruihao~Du,\\    
    Shi Jin, \textit{Fellow, IEEE}, 
    and
    Howard~H.~Yang, \textit{Member, IEEE} 

    \thanks{This paper is supported in part by the National Key R\&D Program of China under Grant 2024YFE0200700 and in part by the National Natural Science Foundation of China under Grant 62201504. 
    (\textit{Corresponding author: Howard H. Yang.})}
		
    \thanks{ J. Zhu, R. Du, and H.~H.~Yang are with the ZJU-UIUC Institute, Zhejiang University, Haining 314400, China (e-mail: \{jiaqi.23, ruihao.24, haoyang\}@intl.zju.edu.cn).}	

    \thanks{ Z. Zhao is with the School of Information and Communication Engineering and the State Key Laboratory of Networking and Switching Technology, Beijing University of Posts and  Telecommunications, Beijing 100876, China (e-mail:  zyzhao@bupt.edu.cn).}

    \thanks{ X. Li and S. Jin are with the School of Information Science and Engineering, Southeast University, Nanjing 210096, China (e-mail: \{li\_xiao, jinshi\}@seu.edu.cn).}

}
\maketitle
\acresetall
\thispagestyle{empty}
\begin{abstract}
Incorporating over-the-air computations (OAC) into the model training process of federated learning (FL) is an effective approach to alleviating the communication bottleneck in FL systems.
Under OAC-FL, every client modulates its intermediate parameters, such as gradient, onto the same set of orthogonal waveforms and simultaneously transmits the radio signal to the edge server.
By exploiting the superposition property of multiple-access channels, the edge server can obtain an automatically aggregated global gradient from the received signal. 
However, the limited number of orthogonal waveforms available in practical systems is fundamentally mismatched with the high dimensionality of modern deep learning models, giving rise to a severe dimension–waveform disparity.
To address this issue, we propose Freshness Freshness-mAgnItude awaRe top-$k$ (FAIR-$k$), an algorithm that selects, in each communication round, the most impactful subset of gradients to be updated over the air.
In essence, FAIR-$k$ combines the complementary strengths of the Round-Robin and Top-$k$ algorithms, striking a delicate balance between \textit{timeliness} (freshness of parameter updates) and \text{importance} (gradient magnitude). 
Leveraging tools from Markov analysis, we characterize the distribution of parameter staleness under FAIR-$k$. 
Building on this, we establish the convergence rate of OAC-FL with FAIR-$k$, which discloses the joint effect of data heterogeneity, channel noise, and parameter staleness on the training efficiency. 
Notably, as opposed to conventional analyses that assume a universal Lipschitz constant across all the clients, our framework adopts a finer-grained model of the data heterogeneity, resulting in a tighter bound to the estimation error.
The analysis demonstrates that since FAIR-$k$ promotes fresh (and fair) parameter updates, it not only accelerates convergence but also enhances communication efficiency by enabling an extended period of local training without significantly affecting overall training efficiency. 
Extensive simulations verify that FAIR-$k$ consistently outperforms several state-of-the-art baselines across diverse FL settings. Furthermore, we implement FAIR-$k$ on a software-defined radio-based prototype, which corroborates its practical effectiveness in real wireless environments.
\end{abstract}
\begin{IEEEkeywords}
Federated learning, over-the-air computing, parameter selection, age of update, convergence rate.
\end{IEEEkeywords}

\acresetall

\section{Introduction}\label{sec:intro} 
Federated learning (FL) has emerged as a paradigm of privacy-preserving distributed machine learning \cite{mcmahan17communication}, but its performance is often constrained by the communication bottleneck, especially when it is operated at the network edge, where communications usually take place over the spectrum \cite{YanLiuQue:20}.
In these cases, integrating over-the-air computations (OAC) \cite{nazer07computation} into the FL model training process stands as an effective solution.
Whereby having all the clients modulate their intermediate parameters, such as gradients, onto a common set of orthogonal waveforms and simultaneously transmit the radio signals to the edge server, OAC enables all the clients to upload their locally trained results in every communication round \cite{zhu19broadband, yang20federated, sery20analog, amiri20federated, chen23over}. 
The edge server then extracts an automatically aggregated gradient from the received signal to update the global model, which is subsequently broadcast back to the clients for further local training. 
Consequently, OAC-FL offers benefits in spectral and energy efficiency, access latency, and privacy protection \cite{elgabli21harnessing, yang21revisiting, yang24unleashing, zhao24model}. 
Despite these advantages, OAC aggregation faces a critical constraint stemming from its reliance on \textit{orthogonal} waveforms, with each waveform typically dedicated to transmitting a single coordinate (entry) of the global gradient.
Indeed, typical communication systems can support up to a hundred thousand orthogonal waveforms (e.g., bases of orthogonal frequency division multiplexing (OFDM) signals \cite{sery20analog}) within the channel coherence time, which is far fewer than the tens of millions of parameters in modern deep neural networks such as ResNet.
This stark mismatch between the scarcity of orthogonal subcarriers in practical wireless systems and the extremely high dimensionality of models creates a dimension–waveform disparity, hindering the implementation of OAC-FL in real-world deployments \cite{amiri20machine}.

In light of the above challenge, we propose Freshness Freshness-mAgnItude awaRe top-$k$ (FAIR-$k$), an algorithm that selects, in each communication round, the most impactful subset of gradients to be updated over the air.
Unlike conventional parameter selection schemes \cite{aji17sparse, stich18sparsified, amiri20machine, ahn22model, zheng25toward} that typically assess importance solely by gradient magnitude, FAIR-$k$ also incorporates the parameter freshness information.
As a result, it strikes a delicate balance between the importance and timeliness in the parameter updates, hence outperforming a string of state-of-the-art methods.
The central goal of this paper is, therefore, to provide a comprehensive introduction to FAIR-$k$, including its design principles, algorithmic structure, and theoretical and empirical performance characterizations.

\subsection{Main Contributions}

The main contributions of this work are summarized below.

\begin{itemize}
    \item We propose an age-aware parameter updating scheme for OAC-FL, referred to as FAIR-$k$. 
    The scheme unites the complementary strengths of the Round-Robin and Top-$k$ algorithms, effectively identifying the most influential subset of gradients to be updated in each round of global communication. 

    \item We establish a theoretical framework to analyze the performance of FAIR-$k$ on OAC-FL model training. Specifically, we derive the convergence rate under FAIR-$k$, which discloses the joint effect of data heterogeneity, channel noise, and parameter staleness on the training efficiency. As a byproduct, we also obtain the parameter staleness distribution under the FAIR-$k$ updating policy, where the accuracy has been verified through simulations. 
   
    \item We validate the effectiveness of FAIR-$k$ through not only empirical simulations, but also a prototype implementation. All the experimental results demonstrate that FAIR-$k$ consistently outperforms several state-of-the-art baselines across diverse FL settings, verifying its effectiveness.
\end{itemize}

\subsection{Prior Works}

\subsubsection{Over-the-Air Federated Learning}
While the OAC-FL efficiently alleviates the communication bottleneck and facilitates highly scalable systems, the inherent channel fading and thermal noise inevitably distort the aggregated gradient.
In response, power control strategies based on instantaneous or statistical channel state information (CSI) estimation have been explored \cite{cao20optimized, cao21optimized, guo22joint, jing22federated} to counteract wireless channel impairments and minimize distortion. 
Recognizing the noise resilience of (stochastic) gradient descent-based algorithms, several studies \cite{sery20analog, yang21revisiting} proposed bypassing explicit CSI estimation and directly utilizing OAC for distributed learning. 
More recently, it has been theoretically revealed that the impact of small-scale fading naturally vanishes as the number of participating clients increases \cite{zhu25rethinking}.

In addition to their detrimental impact on learning performance, channel distortions also help enhance end-user privacy during the training process \cite{liu20privacy, koda20differentially, elgabli21harnessing}.
Despite substantial theoretical and simulation-based advances, only a limited number of works have investigated hardware implementations of OAC-FL \cite{guo21over, csahin22demonstration, pradhan2025experimental}, where the stringent synchronization requirement \cite{xiao24over} presents significant challenges for reliable deployment in practical wireless systems.

\subsubsection{Gradient Sparsification for OAC-FL}

Various gradient compression techniques have been explored in the context of OAC-FL, including sparsification strategies \cite{amiri20machine, li25personalized} and low-rank approximation techniques \cite{xing21federated, makkuva23laser}.
Among them, sparsification, particularly Top-$k$, is widely adopted due to its efficiency by weighting the importance of entries based on magnitude.
However, applying the Top-$k$ sparsification in OAC-FL presents two key challenges.

On the one hand, the positions of the Top-$k$ entries differentiate from each client's local gradient, especially when a large number of clients are present, and the data is highly heterogeneous, making its aggregation not feasible through over-the-air computing.
While \cite{amiri20machine, amiri20federated} investigated the combination of compressed sensing (CS) to reconstruct the aggregated model, the condition for CS to work is often violated in the OAC-FL settings, resulting in noticeable errors in recovery.
Alternatively, \cite{zhang21federated, ahn22model, tao24private} suggested sharing the identical sparsification pattern across the edge server and all clients, to fully utilize the superposition property of wireless waveforms.
However, enforcing a common sparsification mask essentially selects the global Top-$k$ entries, resulting in an inherently inaccurate and less efficient prioritization since local differences are ignored.
                         
On the other hand, due to the temporal correlation of gradient entries across rounds, certain entries may consistently maintain a leading magnitude.
Under aggressive compression, this phenomenon causes many entries to remain unselected for long periods, preventing timely updates and leading to biased or stale model parameters.
To alleviate this, \cite{zheng25toward} introduced a hybrid scheme that combines Top-$k$ with Random-$k$ selection.
The incorporation of Random-$k$ strategy with Top-$k$ have also been investigated in distributed learning \cite{barnes20rtop}.
While injecting randomness helps break the deterministic top selection, effectively balancing the exploration and utilization of gradient information, it still overlooks the explicit timeliness of the parameter update, which is critical for convergence.

\subsection{Notation} 
Throughout the paper, column vectors are represented by bold lowercase letters. 
The L-2 norm of a vector $\boldsymbol{x}$ is denoted by $\|\boldsymbol{x}\|$, and $|\boldsymbol{x}|$ represents obtaining the absolute value of each entry of $\boldsymbol{x}$.
For any positive integer $i$, $[i]$ denotes the set of integers $\{1,2,...,i\}$.
The all-ones vector is denoted by $\boldsymbol{1}$, and $\circ$ denotes the Hadamard product.
The notation $\mathcal{O}(\cdot)$ is used to express an upper bound subsuming universal constants.

\begin{figure*}[t!]
    \centering
    \includegraphics[width=0.985\textwidth]{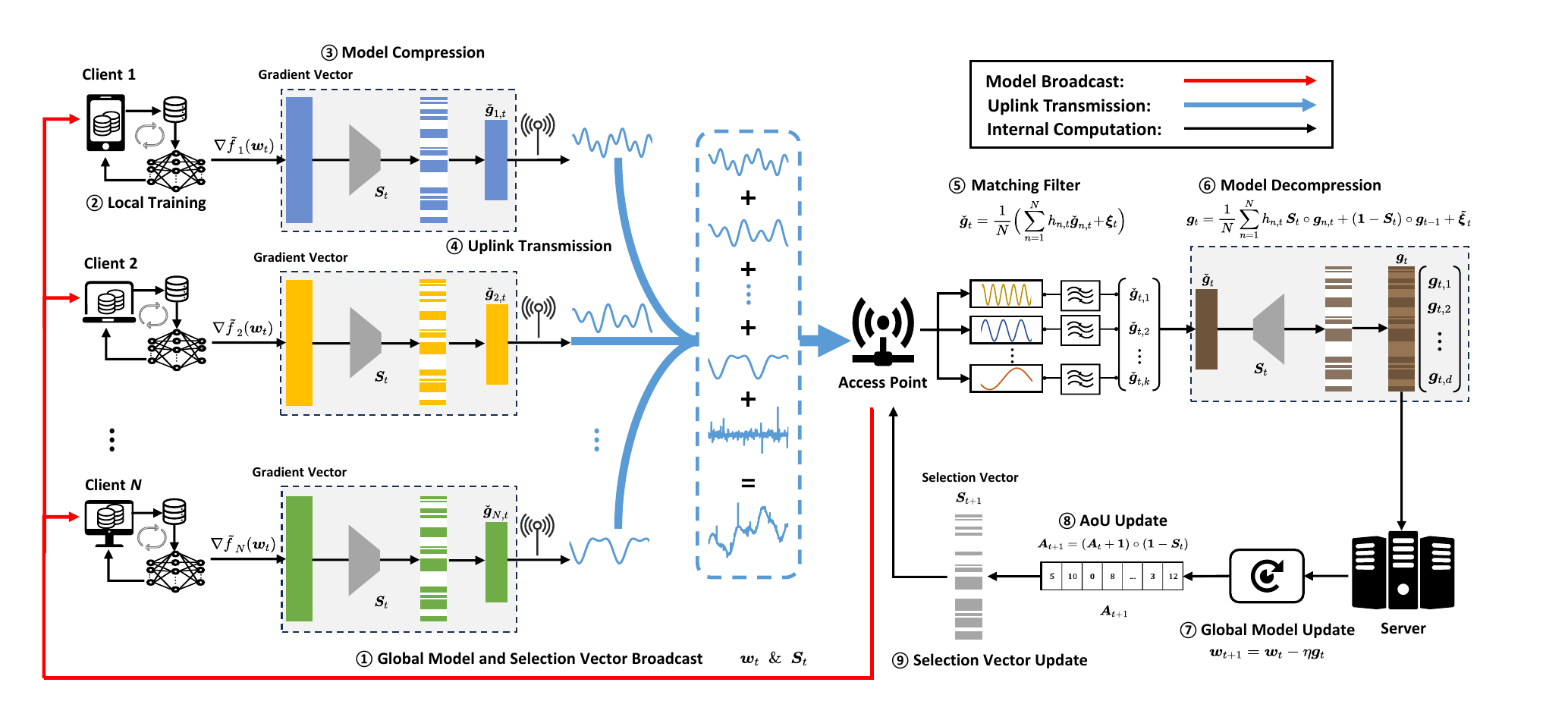} 
    \caption{An overview of the edge learning system. The following steps are repeated until convergence: (1) each client calculates the local gradient based on its local dataset and uploads the compressed gradient to the server via analog transmissions; (2) the server extracts an automatically aggregated global gradient from the received radio signal, and reconstructs it to update the global model; (3) the updated model and selection vector are broadcast to all the clients for a new round of local updating.
    }
\label{fig:system_model}
\end{figure*}

\section{System Model}\label{sec:sysmod}

We consider a federated edge learning system comprised of an edge server and $N$ clients.
Any communications between the clients and the edge server occur over the spectrum.
In this system, each client $n$ owns a local data set $\mathcal{D}_{n} = \{( \boldsymbol{x}_{i}, y_{i} ) \}_{i=1}^{B_n}$, where $\boldsymbol{x}_{i} \in \mathbb{R}^{d}$ is the feature and $y_{i} \in \mathbb{R}$ the label, whilst $B_n$ represents the size of the data set.
We assume the data samples of each client are independent, but do not necessarily follow an identical distribution.

The goal of the edge server is to orchestrate with the clients to train a statistical model that tightly approximates the feature-to-label mapping by leveraging the clients' local data sets while preserving their data privacy.
Formally, the task can be accomplished via the following optimization problem:
\begin{equation}\label{eq:Gbl_loss}
    \min_{\boldsymbol{w}\in \mathbb{R}^{d}} f(\boldsymbol{w})=\frac{1}{N}\sum^{N}_{n=1}f_{n}(\boldsymbol{w}) 
\end{equation}
where $f(\boldsymbol{w})$ is the global loss function and $f_n(\boldsymbol{w})$ is the local loss function of client $n$, constructed by its on-device data set, as:
\begin{equation}
    f_n(\boldsymbol{w})=\frac{1}{B_{n}}\!\sum^{B_{n}}_{i=1}\!\ell(\boldsymbol{w};\boldsymbol{x}_{i},y_{i})
\end{equation}
in which $\ell(\boldsymbol{w};\boldsymbol{x}_{i},y_{i})$ is the loss associated with sample pair $(\boldsymbol{x}_{i},y_{i})$.
The solution of \eqref{eq:Gbl_loss}, commonly known as the empirical risk minimizer, is denoted by 
\begin{equation}
    \boldsymbol{w}^{*}=\arg\min_{ \boldsymbol{w} \in \mathbb{R}^d } f(\boldsymbol{w})
\end{equation}

In this paper, we consider the clients adopt over-the-air federated learning to solve \eqref{eq:Gbl_loss}.
Specifically, during the parameter upload stage of each communication round, the clients modulate their intermediate local parameters (usually the accumulated local gradient) onto the magnitudes of a set of orthogonal waveforms in an entry-wise manner, and simultaneously transmit the resultant radio signal to the edge server, leveraging the superposition property of the multiple access channels for automatic aggregation of the clients' local gradients \cite{chen23edge}.
To accommodate the limited number of available orthogonal carriers, which we denote by $k$, with $k \ll d$, each client updates only a portion of the gradient parameters to the server in each communication round, where the selection of each entry depends on the particular parameter updating policy (which we elaborate in Section~III).
A pictorial example of the model training procedure is shown in Fig.~\ref{fig:system_model} and is also detailed in the next section.

\section{Over-the-Air Model Training }
In this section, we detail the training procedure of OAC-FL systems.
We also elaborate on the design of the FAIR-$k$ algorithm.

\subsection{General Procedure}

At the $t$-th communication round, the edge server broadcasts the global model $\boldsymbol{w}_t$, along with a selection vector $\boldsymbol{S}_t \in \{0,1\}^{d}$ (where $\|\boldsymbol{S}_t\|_1 = k$) to all clients.

Upon receiving the global signal, each client $n$ initializes its local model as $\boldsymbol{w}^{(0)}_{n,t}=\boldsymbol{w}_t$, constructs a set of mini batches of the data samples, $\theta^{(s)}_{n} \in \mathcal{D}_n$, $s\in [H]$, and then executes $H$ steps of local stochastic gradient descent (SGD).
Concretely, at the $s$-th local iteration, client $n$ has its local model update as
\begin{equation}
    \boldsymbol{w}^{(s+1)}_{n,t}=\boldsymbol{w}^{(s)}_{n,t} - \eta_{l}\nabla f_n(\boldsymbol{w}^{(s)}_{n,t};\theta^{(s)}_{n})
\end{equation}
where $\eta_{l}$ denotes the local learning rate.

Once the local training terminates, the client assembles the following accumulated local gradient
\begin{equation} 
    \nabla \widetilde{f}_{n}(\boldsymbol{w}_t)=\sum^{H-1}_{s=0} \nabla f_n(\boldsymbol{w}^{(s)}_{n,t};\theta^{(s)}_{n}).
\end{equation}
Then, by applying an entry-wise filtering of $\nabla \widetilde{f}_{n}(\boldsymbol{w}_t)$ in accordance with $\boldsymbol{S}_t$, the client sparsifies the gradient vector as follows
\begin{align}
    \boldsymbol{g}_{n,t} = \boldsymbol{S}_t \circ \nabla \widetilde{f}_{n}(\boldsymbol{w}_t)
\end{align}
where $\circ$ stands for the Hadamard product.

Note that the vector formed by the non-zero entries of $\boldsymbol{g}_{n,t}$, which we denote by $\check{\boldsymbol{g}}_{n,t}$, has dimension $k$ and hence can be modulated onto the carrier bases.
As such, all the clients simultaneously upload vector $\check{\boldsymbol{g}}_{n,t}$ via analog transmission in the OAC manner.
At the edge server, it receives an automatically aggregated (but distorted) partial gradient vector as follows:
\begin{equation} \label{equ:AccmlGrdt}
    \check{\boldsymbol{g}}_t=\frac{1}{N}\Big(\sum_{n=1}^{N}h_{n,t}\check{\boldsymbol{g}}_{n,t} \!+\! \boldsymbol{\xi}_t\Big)
\end{equation}
where $h_{n,t}$ is the channel fading experienced by client $n$ and $\boldsymbol{\xi}_t \in \mathbb{R}^{k}$ represents the channel noise vector.
We assume the channel fading is independent and identically distributed (i.i.d.) across clients and time, with mean $\mu_c$ and variance $\sigma^2_c$.
Moreover, we consider the noise vector has i.i.d. entries with zero mean and bounded variance $\sigma_z^2$. {\footnote{In practice, the channel noise often exhibits a heavy-tailed distribution, resulting in an excessively large (or even unbounded) variance \cite{yang21revisiting}. In this case, we can apply the clipping technique in \cite{li25robust} for effective denoising.}}

Subsequently, the edge server renews the global gradient with the updated entries. 
The resultant global gradient can be expressed as 
\begin{equation} \label{eq:g_t}
    \boldsymbol{g}_t  = \frac{1}{N} \sum_{n=1}^{N} h_{n,t} \, \boldsymbol{S}_t \circ \boldsymbol{g}_{n,t} + (\boldsymbol{1} - \boldsymbol{S}_t) \circ \boldsymbol{g}_{t-1} + \tilde{\boldsymbol{\xi}}_t
\end{equation}
where $\tilde{\boldsymbol{\xi}}_t$ is constructed from $\boldsymbol{\xi}_t/N$ by padding zeros to the unsent entries. 
Then, the edge server updates the global model as
\begin{equation} \label{equ:OTA_model_training_vanilla}
    \boldsymbol{w}_{t+1}=\boldsymbol{w}_t - \eta\boldsymbol{g}_t
\end{equation}
where $\eta$ denotes the global learning rate.
After that, the system proceeds to the next round of global iteration. 
This process executes recursively until the model converges.

\begin{figure}[!t]
    \label{alg:compress}    
    \begin{algorithm}[H]
	\caption{Freshness-mAgnItude awaRe top-$k$ (FAIR-$k$) gradient compression for OAC-FL}
        \renewcommand{\algorithmicrequire}{\textbf{Input:}}
		\begin{algorithmic}[1]
        \REQUIRE{Initial global model $\boldsymbol{w}_0$, compression dimension $k$, the AoU vector $\boldsymbol{A}_0=\boldsymbol{0}$, and the selection vector $\boldsymbol{S}_{0}=\boldsymbol{1}$}
        \FOR {$t = 0, 1, 2, ..., T\!-\!1$}   
            \STATE Edge server broadcasts $\boldsymbol{w}_t$ and $\boldsymbol{S}_t$ to all the clients  
            \FOR {client $n=1, 2, ..., N$ \textbf{in parallel}}
                \STATEx \textit{\qquad \# Train model locally and update gradients} 
                \STATE $\nabla \widetilde{f}_n(\boldsymbol{w}_t)$ $\leftarrow$ \textsc{ClientUpdate}($n,  \boldsymbol{w}_t, H$) 
                \STATEx \textit{\qquad \# Transmit the compressed gradients} 
                \STATE Apply $\boldsymbol{g}_{n,t} = \boldsymbol{S}_t \circ \nabla \widetilde{f}_{n}(\boldsymbol{w}_t)$, remove the zero entries 
                \STATEx \qquad and transmit the compressed gradient $\check{\boldsymbol{g}}_{n,t}$                
            \ENDFOR
            \STATEx \textit{\quad \# Aggregate via OAC}
            \STATE $\check{\boldsymbol{g}}_{t} = \frac{1}{N}\Big(\! \sum^N_{n=1}h_{n,t}\check{\boldsymbol{g}}_{n,t} \!+\! \boldsymbol{\xi}_t \!\Big)$   
            \STATEx \textit{\quad \# Reconstruct the gradient}
            \STATE $\boldsymbol{g}_t  = \frac{1}{N} \sum_{n=1}^{N} h_{n,t} \, \boldsymbol{S}_t \circ \boldsymbol{g}_{n,t} + (\boldsymbol{1} - \boldsymbol{S}_t) \circ \boldsymbol{g}_{t-1} + \tilde{\boldsymbol{\xi}}_t$
            \STATEx \textit{\quad \# Update global model}
            \STATE $\boldsymbol{w}_{t+1} = \boldsymbol{w}_t - \eta \boldsymbol{g}_t$
            \STATEx \textit{\quad \# Update the AoU}
            \STATE $\boldsymbol{A}_{t+1} = (\boldsymbol{A}_{t} + \boldsymbol{1}) \circ (\boldsymbol{1} - \boldsymbol{S}_t)$
            \STATEx \textit{\quad \# Update the selection vector} 
                \STATE $\boldsymbol{S}_{t+1} = \textsc{SparseSelection}(\boldsymbol{g}_t, \boldsymbol{A}_t, k)$
        \ENDFOR     
	\end{algorithmic}
    \end{algorithm}
\end{figure}

\subsection{Age-Aware Parameter Updating Policy}  
During the above model training process, parameter selection (i.e., $\boldsymbol{S}_t$) plays a central role in determining the training efficiency.
In this part, we detail the FAIR-$k$ algorithm, which strikes a balance between magnitude and freshness of the updated gradient entries. 

Specifically, we leverage the Age of Update (AoU) \cite{yang20age} to capture the freshness of global gradient entries. 
Whereby the edge server maintains a vector $\boldsymbol{A}_t \in \mathbb{R}^d$, that is initialized as $\boldsymbol{A}_{0}=\boldsymbol{0}$ and evolves as \begin{align}
    \boldsymbol{A}_{t+1} = \left( \boldsymbol{A}_t + \boldsymbol{1} \right) \circ \left( \boldsymbol{1} - \boldsymbol{S}_t \right).
\end{align} 
The AoU measures the number of communication rounds since the server last received an updated value for that entry; hence, a larger AoU indicates a higher degree of staleness.

Then, the selection policy of FAIR-$k$ proceeds as follows: (i) sort all entries in descending order of the magnitudes and select the top $k_M$ entries with the largest magnitudes; (ii) among the remaining $d - k_M$ entries, sort them in descending order of their AoU values and select the top $k_A$ entries with the highest AoU values, where $k_A = k - k_M$.
More formally, let $\textsc{Top}(\boldsymbol{x}, k)$ denote an operator that returns a binary vector $\boldsymbol{v}$, with its $i$-th entry $\boldsymbol{v}_{i}=1$ if $\boldsymbol{x}_{i}$ is among the $k$ largest magnitudes of $|\boldsymbol{x}|$, and $\boldsymbol{v}_{i}=0$ otherwise.
The resulting selection vector can be written as
\begin{align}
    \boldsymbol{S}_{t+1} = \textsc{Top}(\boldsymbol{g}_t, k_M) + \textsc{Top}\Big(\! \boldsymbol{A}_t \circ \big(\boldsymbol{1} - \textsc{Top}(\boldsymbol{g}_t, k_M) \big), k_A \!\Big).
\end{align}

The OAC-FL model training under FAIR-$k$ is presented in Algorithm~1.
Note that by supplementing the classical magnitude-based selection with AoU-guided refinement, the FAIR-$k$ scheme exhibits several advantages: (i) it incurs no additional information and maintains low computational complexity, and (ii) it effectively reduces parameter staleness throughout the training process, enabling more balanced and timely updates across model entries.

\begin{remark}
    \textit{We emphasize that when $k_M=k$ (correspondingly, $k_A=0$), FAIR-$k$ becomes Top-$k$.
    Conversely, when $k_M=0$ (correspondingly, $k_A=k$), FAIR-$k$ reduces to the round robin algorithm.   
    FAIR-$k$ therefore unites the complementary strengths of the two methods, being able to identify the most impactful subsets of gradient in each round of local parameter uploading.
    }
\end{remark}

\begin{figure*}[t!]
    \centering
    \subfloat[]
    {
        \includegraphics[width=0.985\linewidth]{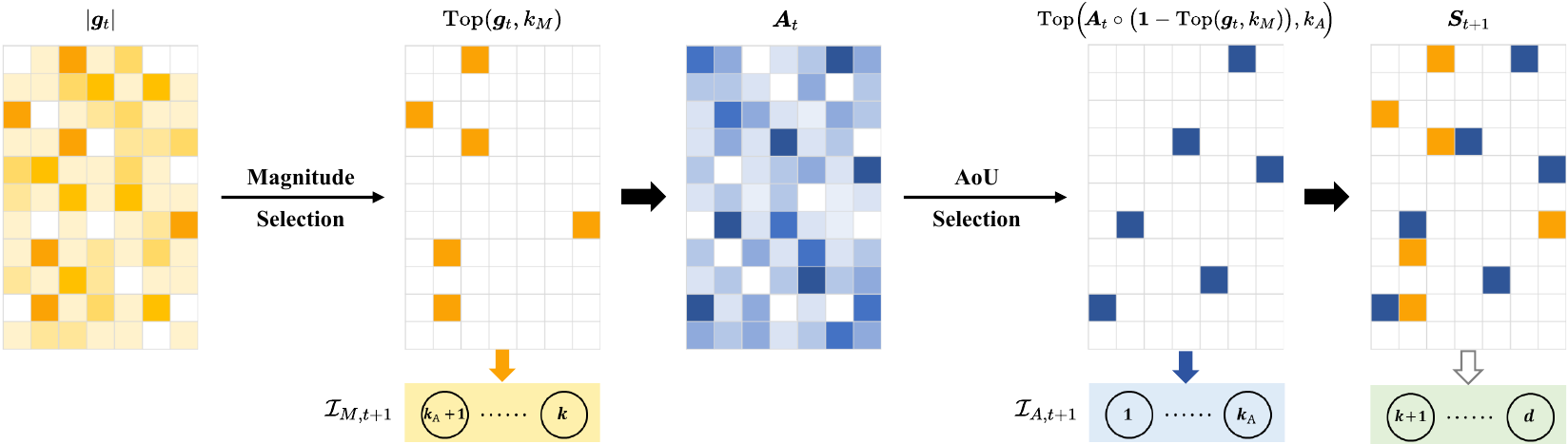}
        \label{fig:selection}
    }
    \\
    \subfloat[]
    {
        \includegraphics[width=0.85\linewidth]{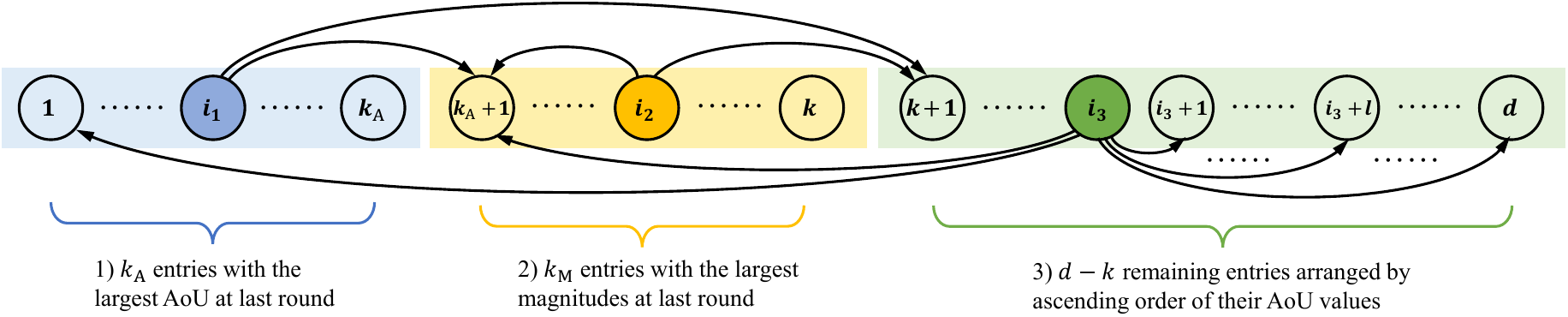}
        \label{fig:chain}
    }
    \caption{Visual illustrations of the selection process for FAIR-$k$ and the dynamics of entry position transition, modeled as a Markov chain.}
    \label{fig:transition}
\end{figure*}


\section{Convergence Analysis}
In this section, we derive the convergence rate of the OAC-FL model training under FAIR-$k$ to examine the efficiency of the proposed algorithm. 
Specifically, since the selection vector $\boldsymbol{S}_t$ varies over the communication rounds, we start by characterizing its dynamic pattern and then use that to derive the convergence rate. 
For better readability, detailed proofs and mathematical derivations are deferred to the appendix.

\subsection{Preliminaries}
Because only a subset of gradient entries is transmitted and incorporated into the model update at each communication round, the remaining entries are not refreshed and thus become stale.
As such, we denote a random variable $\tau_i = \boldsymbol{A}_{t, i}$ as the staleness associated with the $i$-th gradient entry. 
According to \eqref{eq:g_t}, after the $t$-th communication round, the $i$-th entry of the reconstructed gradient at the server side can be rewritten as follows:
\begin{align}
    \boldsymbol{g}_{t,i} = \frac{1}{N} \sum^N_{n=1} h_{n, t-\tau_i} \nabla \widetilde{f}_{n, i} (\boldsymbol{w}_{t-\tau_i}) + \frac{1}{N} \boldsymbol{\xi}_{t-\tau_i, i}.
\end{align}
Note that $\tau_i$ captures how many communication rounds have passed since the $i$-th entry was last updated, and the distribution of $\tau_i$ depends on the parameter selection scheme adopted.
In the sequel, we characterize the dynamics of AoU under FAIR-$k$ by deriving its distribution. 

\subsection{Dynamics of the AoU}
We first define the set of indices selected based on magnitude:
\begin{align}
    \mathcal{I}_{M, t} = \{i: i \in [d], \boldsymbol{v}_{t,i}=1\}
\end{align}
where $\boldsymbol{v}_t = \textsc{Top}(\boldsymbol{g}_t, k_M)$.
Correspondingly, we denote by $\mathcal{I}^c_{M, t}$ the set of unselected indices; within the set $\mathcal{I}^c_{M, t}$, we define the AoU-prioritized subset as:
\begin{align}
    \mathcal{I}_{A, t} = \{i: i \in \mathcal{I}^c_{M, t}, \boldsymbol{v}^A_{t,i}=1\}
\end{align}
where $\boldsymbol{v}^A_t = \textsc{Top}\big( \boldsymbol{A}_t \circ (\boldsymbol{1} - \textsc{Top}(\boldsymbol{g}_t, k_M) ), k_A \big)$.

We model the evolution of elements in $\mathcal{I}_{M,t}$ using a simple exchange process.
Specifically, we assume that at each communication round $t$, the two sets $\mathcal{I}_{M, t}$ and $\mathcal{I}^c_{M, t}$ exchange $k_0$ entries, where $k_0 < k_M$.
To keep the analysis tractable, we assume that the parameter exchange occurs uniformly at random across all entries.
Concretely, the transition probabilities for a typical entry $i$ are 
\begin{align}
    \begin{cases}
    \mathbb{P}(i \in \mathcal{I}^c_{M, t+1} | i \in \mathcal{I}_{M, t})= p_1,  \\
    \mathbb{P}(i \in \mathcal{I}_{M, t+1} | i \in \mathcal{I}_{M, t})= 1-p_1, \\
    \mathbb{P}(i \in \mathcal{I}_{M, t+1} | i \in \mathcal{I}^c_{M, t})= p_2, \\
    \mathbb{P}(i \in \mathcal{I}^c_{M, t+1} | i \in \mathcal{I}^c_{M, t})= 1-p_2  \\
    \end{cases}
\end{align}
where we stipulate that $p_1=\frac{k_0}{k_M}$ and $p_2=\frac{k_0}{d-k_M}$.
The transition probabilities describe how likely a given entry is to be selected or remain unchanged across communication rounds, thereby capturing the temporal correlation inherent in the evolution of gradient entries.
Although simplified, the model captures an essential property of practical distributed training, that when $k_0 < \frac{k_M(d-k_M)}{d}$, we have $1-p_1 > p_2$, indicating that entries selected by the Top-$k_M$ algorithm are more likely to remain in $\mathcal{I}_{M, t}$ compared to entries originally in $\mathcal{I}^c_{M,t}$.
This aligns with the empirical observation that large-magnitude gradient entries tend to persist across consecutive rounds.

Next, we derive the distribution of parameter staleness.
To facilitate analysis, we arrange the entries of the global gradient in ascending order of their AoU values at each communication round.
As a result, the first entry has the smallest AoU (which corresponds to the freshest information), while the last entry has the largest AoU (meaning it is the most stale).
As training proceeds, the position of a generic entry $i$ varies according to the AoU vector.
Fig.~\ref{fig:transition} depicts the possible transition states of a typical entry; aided by this figure, we detail the subsequent analysis below.

At round $t$, an entry resets its AoU to zero if (i) it is categorized into the top-magnitude set $i \in \mathcal{I}_{M, t}$, or (ii) it belongs to $i \in \mathcal{I}^c_{M, t}$ and falls within the top $k_A$ entries with the largest AoU values.
In total, $k$ entries satisfy $\boldsymbol{A}_{t,i} = 0$.
For ease of exposition, we assign states $1$ to $k_A$ to the AoU-prioritized entries in the set $\mathcal{I}_{A,t}$, and $k_A+1$ to $k$ to the magnitude-prioritized entries in the set $\mathcal{I}_{M,t}$ (as shown in Fig.~\ref{fig:selection}).
Note that the entries within $\mathcal{I}_{A,t}$ and $\mathcal{I}_{M,t}$ are not ordered, and we simply denote all the positions in the sets by their first position, i.e., position $1$ for $\mathcal{I}_{A,t}$ and position $k_A +1$ for $\mathcal{I}_{M,t}$, respectively.  
In this paper, we only consider the regime where the compression ratio is at most $\rho \le 50\%$, i.e., $k \le \frac{d}{2}$, which is relevant to practical scenarios.

We model the clients' positions as the states of a Markov chain, with the transition matrix given by $\mathbf{P}=[\mathbf{P}_{i,j}]_{1\leq i,j \leq d}$.
For a generic entry $i$, its position can experience the following transitions after one global iteration:
\begin{itemize}
    \item If $i \leq k_A$, it belongs to $i_1$ in Fig.~\ref{fig:chain}, there are two possible state transitions:
    \begin{enumerate}
        \item $i \rightarrow k_A+1$: If the entry is selected by Top-$k_M$, this occurs with probability $\mathbf{P}_{i,k_A+1} = p_2$.
        \item $i \rightarrow k+1$: If the entry is excluded from Top-$k_M$, this happens with probability $\mathbf{P}_{i,k+1} = 1 - p_2$.
    \end{enumerate}

    \item If $k_A+1 \leq i \leq k$, it belongs to those $i_2$ in Fig.~\ref{fig:chain}, where there are two possible state transitions:
    \begin{enumerate}
        \item $i \rightarrow k_A+1$: If the entry remains within the top $k_M$ entries with the largest magnitude, this happens with probability $\mathbf{P}_{i,k_A+1} = 1 - p_1$.
        \item $i \rightarrow k+1$: If it is no longer selected by Top-$k_M$, this occurs with probability $\mathbf{P}_{i,k+1} = p_1$.
    \end{enumerate}

    \item If $i \geq k+1$, it falls in $i_3$ Fig.~\ref{fig:chain}, there are three possible state transition:
    \begin{enumerate}
        \item $i \rightarrow k_A+1$: If the entry is selected by Top-$k_M$, this yields the probability $\mathbf{P}_{i,k_A+1} = p_2$.
        \item $i \rightarrow i+k_A+\ell$: If the entry is not selected by Top-$k_M$, and $\ell$ entries with a larger AoU than $i$ are selected by Top-$k_M$; this results in a probability of $\mathbf{P}_{i,i+k_A+\ell} = (1-p_2) \binom{d-i}{\ell}p_2^\ell (1-p_2)^{d-i-\ell}$.{\footnote{Note that the step length $\ell$ is strictly limited ($\ell \le \min\{k_0, d-i\}$), we apply a normalization procedure to the calculated probabilities to ensure the sum over the restricted range equals one.}}
        \item $i \rightarrow 1$: If the entry is not selected by Top-$k_M$, but it subsequently becomes one of the $k_A$ oldest entries after the Top-$k_M$ selection, it is then selected by the AoU prioritization with its AoU reset; this happens with probability $\mathbf{P}_{i,1} = (1-p_2) \sum^{d-i}_{\ell=d-i-k_A} \binom{d-i}{\ell}p_2^\ell (1-p_2)^{d-i-\ell}$.
    \end{enumerate}
\end{itemize}

As a result, the Markov chain is recurrent and irreducible and hence has a steady-state distribution. 
Let $\boldsymbol{\pi}=(\pi_1, \pi_2, ..., \pi_d)$ denote the steady state probability vector; then we can solve for the value of each entry via the following fixed-point equation:
\begin{align}
    \boldsymbol{\pi} = \boldsymbol{\pi} \mathbf{P}.
\end{align}
Following the above transition process, the random variables $\{\tau_i\}_{i=1}^{d}$ become i.i.d.; we denote by $\tau \stackrel{\text{d}}{=} \tau_i$, $\forall i \in [N]$ (where $\stackrel{\text{d}}{=}$ stands for equal in distribution), and characterize its distribution by the following.

\begin{lemma}
    \textit{The AoU distribution is given by  
    \begin{align}
        \mathbb{P}(\tau = l) = \!\sum^d_{i=1} \pi_i \!\left(\!\!\left( \mathbf{P}^l_{(1, k_A+1)} \mathbf{P}\right)_{i, 1} \!+\! \left( \mathbf{P}^l_{(1, k_A+1)} \mathbf{P}\right)_{i, k_A+1} \right)
    \end{align}
    where $l \in [\mathcal{T}]$ with $\mathcal{T} = \frac{d-k_M}{k_A}$ representing the maximum staleness, $\mathbf{P}_{(1,k_A+1)}$ is a matrix obtained by replacing the first and $(k_A+1)$-th column of $\mathbf{P}$ with all zeros, respectively, and $(\mathbf{X})_{i,j}$ denotes the entry $(i,j)$ of matrix $\mathbf{X}$.    
    }
\end{lemma}
\begin{IEEEproof}
    Please see Appendix~A.     
\end{IEEEproof}

\begin{figure}[t!]
    \centering
    \includegraphics[width=0.985\linewidth]{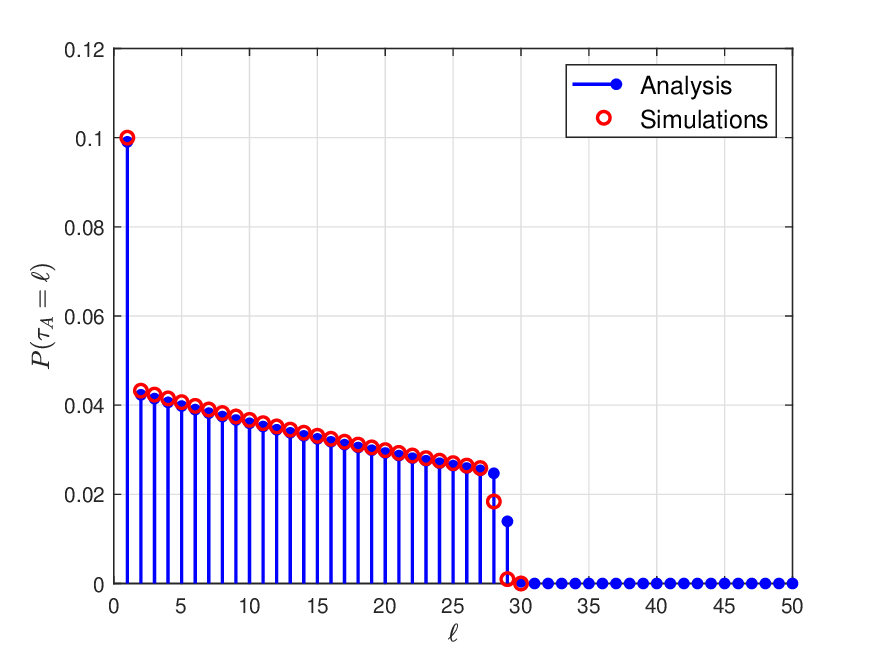}
    \caption{The distribution of AoU.}
    \label{fig:AoU}
\end{figure}

In Fig.~\ref{fig:AoU}, we plot the simulation results to verify the accuracy of our analysis for the AoU distribution. 
We use the following parameters for the number of selected entries, compression ratio, ratio of magnitude selection, and ratio of entry exchange, respectively: $k=80$, $\rho=0.1$, $k_M/k=0.75$, and $k_0/k_M=0.25$.
The figure shows a close match between the simulations and the analytical results, confirming the accuracy of Lemma~1.

\subsection{Convergence Rate}
First, we employ the following \cite{wang24new} to account for data heterogeneity. 

\begin{assumption}\label{assm:glb-L}
    \textit{The global objective function $f: \mathbb{R}^{d} \rightarrow \mathbb{R}$ satisfies the following:
    \begin{align}
        \|\nabla f(\boldsymbol{w}) - \nabla f(\boldsymbol{v}) \| \leq L_g \| \boldsymbol{w}- \boldsymbol{v} \|, \qquad \forall \boldsymbol{w}, \boldsymbol{v} \in \mathbb{R}^{d}
    \end{align}
    where $L_g$ is a positive constant.
    }
\end{assumption}

\begin{assumption}\label{assm:local-L}
    \textit{There exists a constant $L_h \geq 0$ such that for any $\boldsymbol{w}_n \in \mathbb{R}^{d}$, the following is satisfied:
    \begin{align} \label{equ:Pseudo-Lipschitz}
        \Big\|\frac{1}{N} \sum^N_{n=1} \nabla f_n(\boldsymbol{w}_n) - \nabla f(\bar{\boldsymbol{w}}) \Big\|^2 \leq \frac{L_h^2}{N} \sum^N_{n=1} \|\boldsymbol{w}_n - \bar{\boldsymbol{w}} \|^2
    \end{align}
    where $\bar{\boldsymbol{w}} = \frac{1}{N}\sum^N_{n=1}\boldsymbol{w}_n$.
    }
\end{assumption}

It is important to note that these two assumptions provide a fine-grained characterization of the data heterogeneity.
As opposed to the conventional smoothness condition \cite{yu19parallel, khaled20tighter} where all the clients have their local loss functions abide by a common Lipschitz constant, i.e., $\Vert \nabla f_{n}( \boldsymbol{w} ) -\nabla f_{n}( \boldsymbol{v} ) \Vert \leq \tilde{L} \Vert  \boldsymbol{w} -  \boldsymbol{v} \Vert $, $\forall n \in [N]$, Assumption~1 imposes a Lipschitz constant only on the global loss function.
In addition, the constant $L_h$ in \eqref{equ:Pseudo-Lipschitz}, referred to as the \textit{heterogeneity-driven pseudo-Lipschitz constant}, characterizes the difference between the average model and central model, reflecting the actual impact of data heterogeneity in the system.
As will be demonstrated in Section~\ref{sec:experiments}, for a variety of machine learning tasks, $L_g$ and $L_h$ are much smaller than $\tilde{L}$, hence providing a tighter estimation of the convergence rate.

To facilitate the analysis, we make the following additional assumptions.
\begin{assumption}\label{assm:SGD noise}
    \textit{For every client $n$, the stochastic gradient $\nabla f_n(\boldsymbol{w};\theta_{n})$ calculated based on a mini-batch $\theta_{n}$ is an unbiased estimation of $\nabla{f}_{n}(\boldsymbol{w})$ with bounded variance, i.e.,    
    \begin{align}
        &\mathbb{E} \left[ \nabla f_n(\boldsymbol{w};\theta_{n}) \right] = \nabla{f}_{n}(\boldsymbol{w}),\\
        &\mathbb{E} \left[ \|\nabla f_n(\boldsymbol{w};\theta_{n})-\nabla{f}_{n}(\boldsymbol{w})\|^2 \right]
        \leq \sigma^2_{s}.
    \end{align}    
    }
\end{assumption} 

\begin{assumption}\label{assm:gradient bound}
    \textit{The expected squared norm of stochastic gradients $f_n(\boldsymbol{w};\theta_{n})$ is bounded, i.e., for any $n\in[N]$, there exists a positive constant $G$ that
    \begin{equation}
        \mathbb{E}[\|\nabla f_n(\boldsymbol{w};\theta_{n})\|^2] \leq G^2.
    \end{equation}
    }
\end{assumption}

\begin{assumption}\label{assm:hete-bound}
    \textit{There exists a constant $\sigma_g > 0$ such that  
    \begin{equation}
        \|\nabla f_n(\boldsymbol{w})-\nabla f(\boldsymbol{w})\|^2 \leq \sigma_g^2.
    \end{equation}
    }    
\end{assumption}

Note that Assumption~3 is standard in stochastic optimization, and Assumption~4 is generally valid in OAC systems \cite{sery21over} due to the maximum transmit power constraint, where excessively large parameter entries must be trimmed before transmission.
Moreover, Assumption~5 is widely known as the \textit{gradient divergence}, which represents data heterogeneity from a gradient perspective; an increase in $\sigma_g$ indicates a higher degree of heterogeneity across local datasets \cite{li19convergence}.

Subsequently, we establish the following lemma to characterize the upper bound of the reconstructed gradient.
\begin{lemma}
    \textit{Across all the communication round $t$, the following holds:
    \begin{align}
        \mathbb{E}\!\left[ \|\boldsymbol{g}_t\|^2\right]
        &\leq 2(\mu_c^2 \!+\! \sigma_c^2) \!\sum^{\mathcal{T}}_{l=0} q_l \mathbb{E} \Bigg[ \bigg\| \frac{1}{N} \!\sum^N_{n=1} \!\sum^{H-1}_{s=0}\! \nabla {f}_{n} (\boldsymbol{w}_{n,t-l}^{(s)})  \bigg\|^2\Bigg]
        \nonumber \\
        &\quad  +  \frac{2H \sigma_s^2 (\mu_c^2 \!+\! \sigma_c^2)}{N} + \frac{d \sigma_z^2}{N^2}
    \end{align} 
    where $q_l = \mathbb{P}(\tau = l)$, with $l \in [\mathcal{T}]$.
    }
\end{lemma}
\begin{IEEEproof}
     Please see Appendix~B.
\end{IEEEproof}

After the above preparation, we are now ready to present the convergence rate.

\begin{theorem}\label{theo:ConRate}
    \textit{Setting the global and local learning rates as $\eta \leq \frac{\mu_c}{2H L_g (\mu_c^2 + \sigma_c^2)}$ and $\eta_l \leq \min \big\{ \frac{1}{2 \sqrt{30}H L_g}, \frac{1}{\sqrt{6H(L_g^2+L_h^2)}}\big\}$, respectively, model training under the FAIR-$k$ algorithm will converge as    
    \begin{align} \label{equ:CnvgRt_FixedLrnRt}
        &\min_{0 \leq t \leq T-1} \mathbb{E}[\|\nabla f(\boldsymbol{w}_{t})\|^2] = \mathcal{O}\Bigg( \frac{f(\boldsymbol{w}_0) \!-\! f(\boldsymbol{w}^*)}{\eta \mu_c H T}
        \!+\!\frac{\eta d L_g \sigma_z^2}{\mu_c H N^2}
        \nonumber \\
        &\!+\!\frac{\eta L_g \sigma_s^2 (\mu_c^2 \!+\! \sigma_c^2)}{\mu_c N}
        \!+\! (H\!\!-\!\!1)^2 \eta_l^2 L_h^2 \sigma_g^2 \!+\!  (H\!\!-\!\!1)\eta_l^2 \sigma_s^2 \!\bigg(\!\! L_h^2 \!+\! \frac{L_g^2}{N} \!\bigg)            
        \nonumber \\
        &\!+\! \frac{\eta L_g  \mathbb{E}[\tau]}{H}\!\left(\!\frac{d \sigma_z^2}{N^2} \!+\! G^2 H^2(1 \!+\! \mu_c^2 \!+\! \sigma_c^2)\!\right)\!\!\! \Bigg).
    \end{align}  
    }
\end{theorem}
\begin{IEEEproof}
    Please see Appendix~C.
\end{IEEEproof}

This result reveals the interplay among data heterogeneity, multi-step local SGD, communication noise, and parameter staleness (determined by the updating policy) on the convergence rate.
Particularly, the first term on the right-hand side of \eqref{equ:CnvgRt_FixedLrnRt} shows that the global model converges at the rate of $\mathcal{O}(\frac{1}{T})$, whereby adequately scheduling the global and local learning rates, the residual error, which corresponds to the rest on right-hand side of \eqref{equ:CnvgRt_FixedLrnRt}, can be confined with a desire level.
Indeed, OAC-FL is most effective in large-scale systems \cite{zhu25rethinking} where $N \gg 1$. 
Under such circumstances, the convergence rate (asymptotically) reduces to the following
    \begin{align} 
        &\min_{0 \leq t \leq T-1} \mathbb{E}[\|\nabla f(\boldsymbol{w}_{t})\|^2] = \mathcal{O}\Bigg( \frac{f(\boldsymbol{w}_0) \!-\! f(\boldsymbol{w}^*)}{\eta \mu_c H T}
        \nonumber \\
        &\quad + \eta_l^2 (H\! - \!1)^2 L_h^2 ( \sigma_g^2 +\sigma_s^2 ) + \eta H L_g  \mathbb{E}[\tau]G^2 (1 \!+\! \mu_c^2 \!+\! \sigma_c^2) \Bigg)
    \end{align} 
from which we clearly see that by appropriately choosing $\eta_l$ and $\eta$, the residual error can be arbitrarily small.
Therefore, the global model eventually lands in a noisy ball centered at a local minima, with a controllable radius. (Note that since the training process is based on SGD, it effectively avoids saddle points \cite{jin21nonconvex}.)

Moreover, we note that when $H=1$, i.e., the system executes FedSGD, the local gradient deviation, $(H-1)^2 \eta_l^2 L_h^2 \sigma_g^2 +  (H-1)\eta_l^2 \sigma_s^2 ( L_h^2 + {L_g^2}/{N} ) $, vanishes. 
In contrast, if we extend the local iteration steps, $H$, the local gradient deviations increase. 
Nevertheless, because $L_h$ and $L_g$ are normally small, the local gradient deviation goes up at a much lower rate compared to conventional analysis that adopts $\tilde{L}$.
This indicates that the system can tolerate much larger local iterations, enabling infrequent global communications. 

Finally, the last term of \eqref{equ:CnvgRt_FixedLrnRt} displays the joint effect of parameter staleness, local iterations, and communication noise on the convergence rate. 
It shows that the updating policy that results in small parameter staleness also leads to fast training convergence.

\begin{figure*}[t!]
    \centering
    \subfloat[]
    {
        \includegraphics[width=0.44\linewidth]    {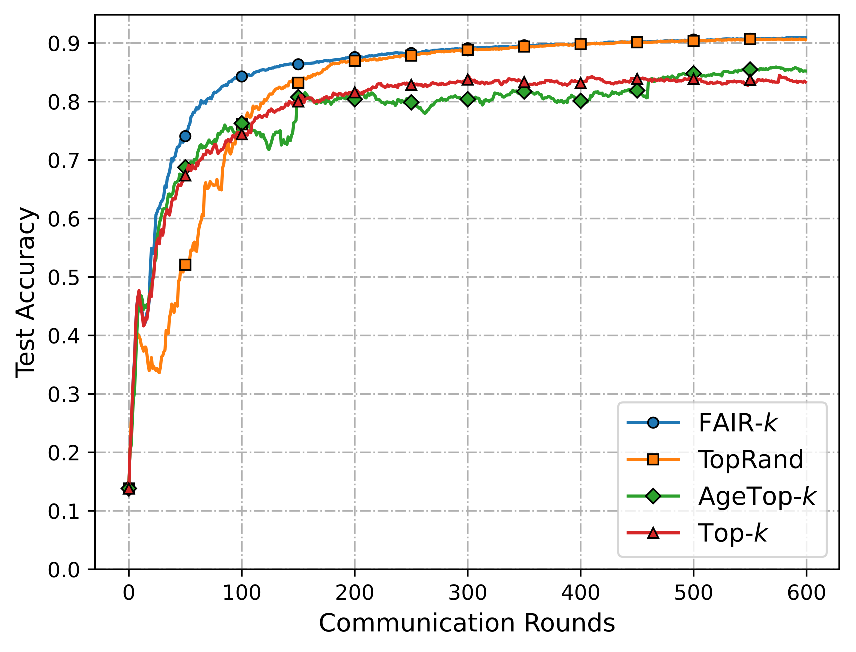}
        \label{fig:subfig_a}
    }
    \hfil
    \subfloat[]
    {
        \includegraphics[width=0.44\linewidth]    {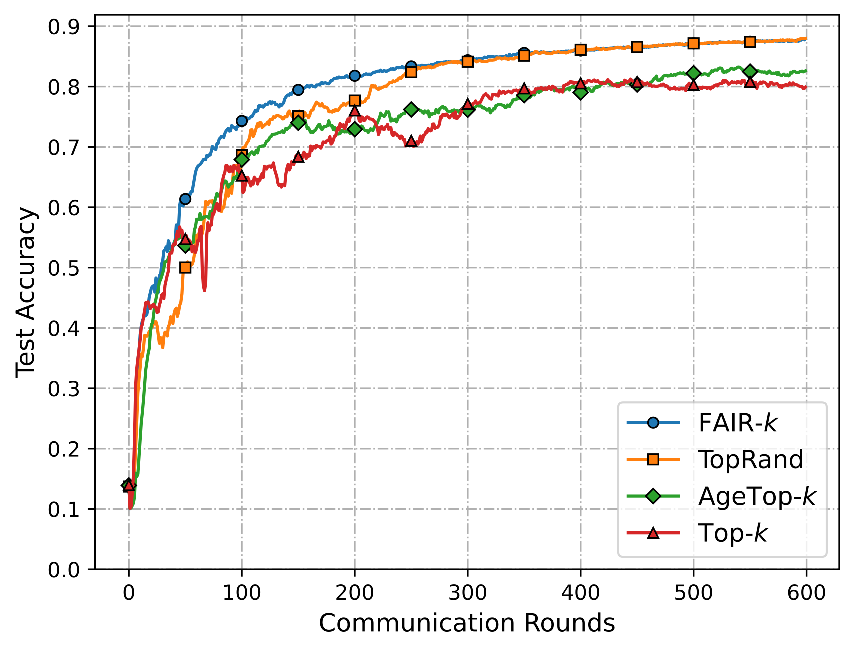}
        \label{fig:subfig_b}
    } 
    
    \subfloat[]
    {
        \includegraphics[width=0.44\linewidth]    {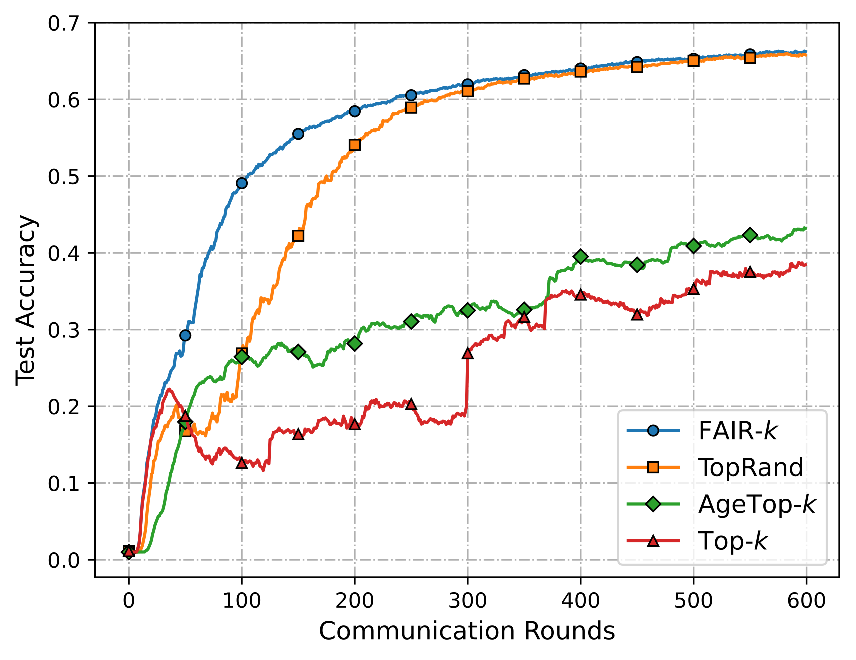}
        \label{fig:subfig_c}
    }
    \hfil
    \subfloat[]
    {
        \includegraphics[width=0.44\linewidth]    {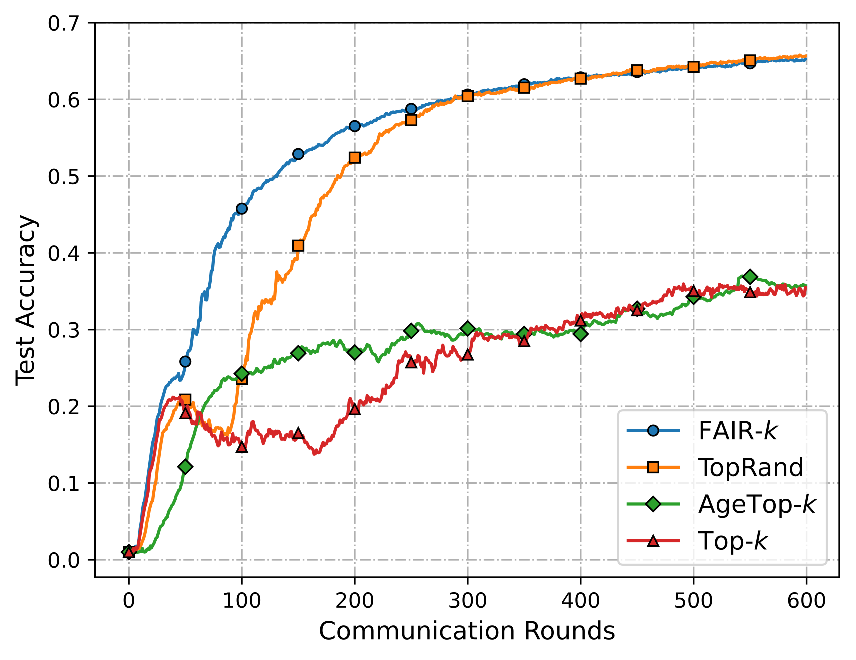}
        \label{fig:subfig_d}
    }
    \caption{Performance comparison for test accuracy.
    Here, (a) and (b) result from training ResNet-18 on the CIFAR-10 dataset with i.i.d. and non-i.i.d. partitions, respectively, while (c) and (d) result from training ResNet-18 on the CIFAR-100 dataset with i.i.d. and non-i.i.d. partitions, respectively.}
    \label{fig:comparison}
\end{figure*}

\section{Experiments} \label{sec:experiments}

In this section, we examine the performance of the proposed FAIR-$k$ scheme through several empirical simulations and a prototype implementation. 
Specifically, we first present the simulation setup and compare FAIR-$k$ with several baselines to demonstrate its efficacy.
Then, we implement the OAC-FL system with FAIR-$k$ through a prototype under real-world wireless conditions.  
We detail the configuration of the prototype, and compare FAIR-$k$ with the same set of baselines as in the simulations to corroborate its practical viability.

\subsection{Empirical Simulations}

\subsubsection{Setup}

We carry out simulations on image classification tasks based on CIFAR-10 and CIFAR-100 datasets \cite{krizhevsky09learning} using ResNet-18 \cite{he16deepresidual}, which consists of a series of convolutional layers and residual connections.
The CIFAR-10 dataset comprises 60,000 images across 10 classes, and the CIFAR-100 is partitioned into 100 classes, with each class holding 600 images. 
Both datasets are divided into a training set of 50,000 images and a test set of 10,000 images.

Unless otherwise specified, the wireless channel gain is modeled as Rayleigh fading with mean $\mu_c=1$. The thermal noise is modeled as additive white Gaussian noise with unit variance, namely, $\boldsymbol{\xi}_t \sim \mathcal{N}(\boldsymbol{0}, \boldsymbol{I})$.
The training set is distributed across $N=50$ clients in a non-i.i.d. manner, with heterogeneity in both class distributions and local dataset sizes. 
Specifically, we adopt the symmetric Dirichlet partitioning \cite{hsu19measuring}, where the heterogeneity level is controlled by the parameter $Dir$, set to $0.3$ in our experiments.
Additionally, we set the local and global learning rates as $\eta_l = \eta= 0.01$, with local batch size and epochs being $B=50$ and $H=5$, respectively.
We denote the compression ratio by $\rho = k / d$.
All experiments are implemented with Pytorch on an NVIDIA RTX 3090 GPU.

\subsubsection{Performance evaluation}
We start by contrasting the training performance under FAIR-$k$ to three baseline methods, i.e., the Top-$k$, AgeTop-$k$ \cite{du25age}, and TopRand \cite{zheng25toward}, with a compression ratio of $\rho=10\%$.
We set $k_M=0.75k$ for both FAIR-$k$ and TopRand, and set $r=1.5k$ for AgeTop-$k$.

In Fig.~\ref{fig:comparison}, we draw the test accuracy as a function of communication rounds under different parameter updating policies. 
From this figure, we can see that FAIR-$k$ consistently demonstrates the fastest convergence rate against the baselines, across different tasks (i.e., image classification on CIFAR-10 and CIFAR-100) and data heterogeneity levels (i.i.d. and non-i.i.d.).
Notably, the performance gain from FAIR-$k$ is especially pronounced in executing complex tasks such as training ResNet-18 on the CIFAR-100 data set. 
Compared to the widely used Top-$k$ algorithm (or its age-based variant AgeTop-$k$), FAIR-$k$ boosts the test accuracy by more than 30\% with minimal cost in extra computations and/or side information (that is, the AoU, which can be obtained without additional communication overheads from the clients).
Even compared to TopRand, which, similarly, decomposes Top-$k$ into a two-stage selection algorithm, FAIR-$k$ significantly accelerates the training speed.
The main reason for such a gain in performance can be attributed to the intrinsic balance FAIR-$k$ strikes between the magnitude and freshness of gradient information.

\begin{figure}[t!]  
    \centering 
    \begin{subfigure}{0.985\linewidth}
		\includegraphics[width=\linewidth]{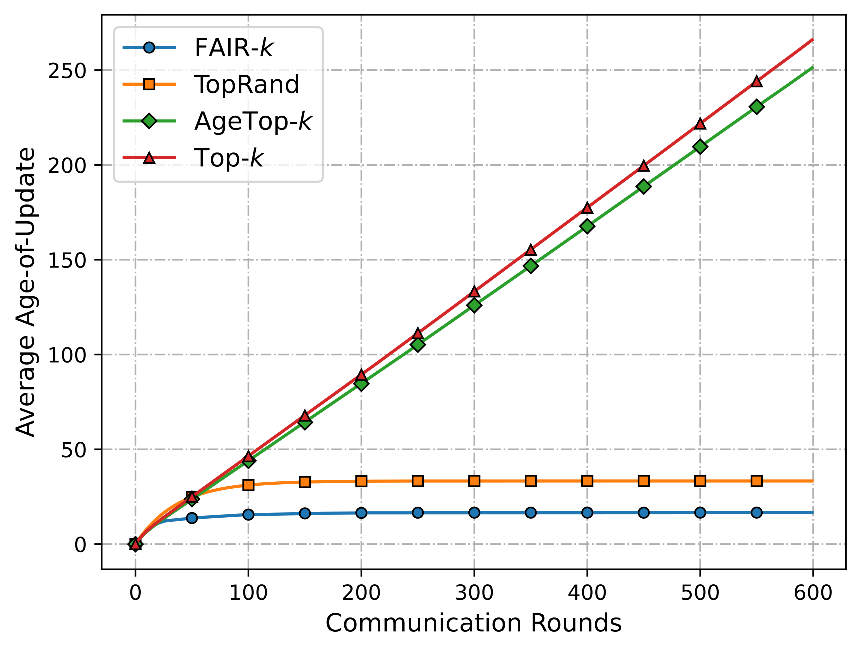}
        \caption{Comparison for average AoU}
        \label{fig:age_comparison}
    \end{subfigure}
    \begin{subfigure}{0.985\linewidth}
		\includegraphics[width=\linewidth]{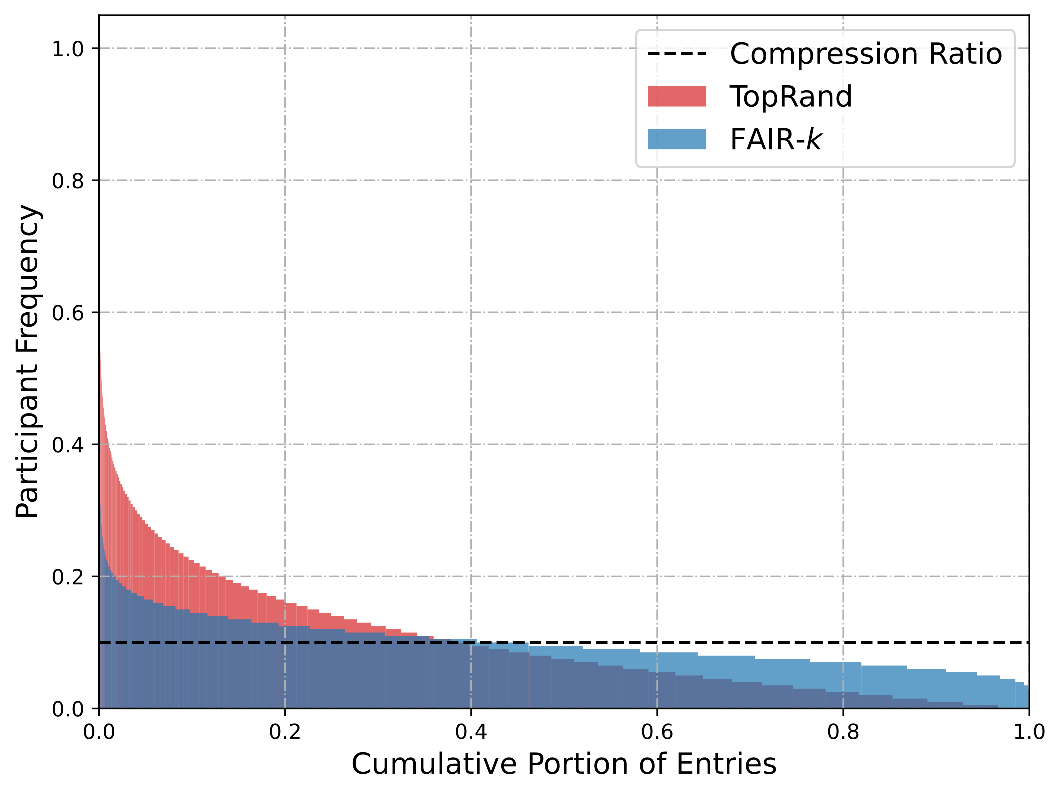}
        \caption{Entry participant frequency in 200 rounds}
        \label{fig:frq_comparison}
    \end{subfigure}
    \caption{The impact of parameter updating schemes on parameter staleness, exemplified by training ResNet-18 on the CIFAR-10 dataset.}
    \label{fig:staleness}
\end{figure}

To further demystify the effectiveness of FAIR-$k$, we display the statistics of AoU and partition frequency in Fig.~\ref{fig:staleness}.
From Fig.~\ref{fig:age_comparison}, we note that FAIR-$k$ achieves the lowest average AoU among all the schemes. 
According to Theorem~\ref{theo:ConRate}, this yields the fastest convergence rate (since the other system factors remain unchanged). 
In addition, Fig.~\ref{fig:frq_comparison} reports the selection frequency of each entry after 200 communication rounds. 
The findings confirm that FAIR-$k$ expands the subset of entries that receive a fairer opportunity to participate in model updates, while still preserving priority for those with large magnitudes.
By first selecting an initial subset of entries based on magnitude, followed by AoU-driven refinement (or random sampling for TopRand) over the remaining entries, both FAIR-$k$ and TopRand are able to effectively cap the growth of average AoU once it reaches a certain threshold.
Benefiting from its age-aware mechanism, FAIR-$k$ reduces the average AoU to nearly half that of the TopRand approach.

\begin{figure}[t!]  
    \centering 
    \begin{subfigure}{0.985\linewidth}
		\includegraphics[width=\linewidth]{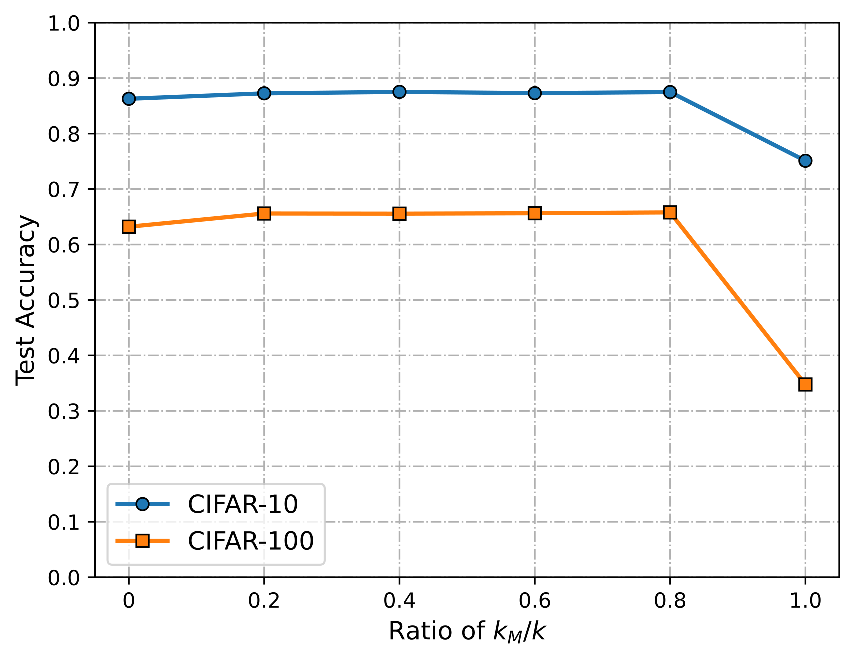}
    \end{subfigure}
    \caption{The impact of the ratio $k_M/k$ on test accuracy, exemplified by training ResNet-18 on the CIFAR-10 and CIFAR-100 dataset, respectively, after 600 communication rounds.}
    \label{fig:k_M}
\end{figure}

\subsubsection{Effects of system factors}
In Fig.~\ref{fig:k_M}, we plot the test accuracy obtained from running OAC-FL with FAIR-$k$ for 600 communication rounds as a function of the ratio $k_M/k$. 
By varying this ratio, we aim to explore the optimal configuration of FAIR-$k$ (based on specific tasks employed). 
Notably, when $k_M=k$, FAIR-$k$ reduces to Top-$k$, while in the case of  $k_M=0$, it becomes round robin.
From this figure, we observe that for a wide range of $k_M/k$, the test accuracy of FAIR-$k$ remains stable, indicating that FAIR-$k$ does not require a delicate configuration of its parameters to produce an effective performance boost.

Additionally, we explore the benefits of FAIR-$k$ in terms of communication efficiency in Table~\ref{tbl:L} and Fig.~\ref{fig:local epoch}.

In particular, we enlist the empirical estimation of Lipschitz constants $\tilde{L}$, $L_g$, and $L_h$ in a variety of machine learning tasks in Table~\ref{tbl:L}.
The table shows that for the same task, $\tilde{L}$, evaluated under the conventional smoothness assumption, can be orders of magnitude larger than $L_g$ and $L_h$, the refined estimates of heterogeneity discrepancy across clients' local data sets. 
Moreover, $\tilde{L}$ is significantly influenced by the degree of data heterogeneity; as the Dirichlet parameter $Dir$ decreases (implying stronger non-i.i.d. distributions), the value of $\tilde{L}$ increases sharply.
Consequently, the conventional convergence rate represents a pessimistic assessment of the increase in discrepancy caused by multiple local epochs.  
By contrast, the convergence rate given in Theorem~\ref{theo:ConRate} implies that the local SGD iterations can be substantially expanded without significantly affecting the training efficiency. 

Indeed, we summarize the model training result with an increasing number of local iterations in Fig.~\ref{fig:local epoch}.
From this figure, we can see that by increasing the local epochs from $H=1$ (which corresponds to the FedSGD setting) and $H=5$ (which is the commonly adopted value in FL) to $H=20$, a relatively long local training period, the model training under OAC-FL is able to converge, regardless of the parameter updating scheme employed.
Moreover, FAIR-$k$ outperforms Top-$k$ by not only enhancing the convergence rate, but also reducing fluctuations in the training process, thus improving system stability. 


\begin{table}[!t]
    \caption{\centering{ Estimation of Lipschitz constants}}
    \label{tbl:L}
    \tabcolsep=4pt    
    \renewcommand\arraystretch{1.5}
    \centering
    \begin{tabular}{c c c c c}
    \toprule
        Dataset & Non-i.i.d Degree & $\tilde{L}^2$ & $L_g^2$ & $L_h^2$  \\ 
    \hline
        \multirow{4}{*}{CIFAR-10} & $Dir=0.1$ & 3193.38 & 159.77 & 8.89 \\ 
         & $Dir=0.3$ & 1198.54 & 108.58 & 5.66 \\ 
         & $Dir=0.5$ & 730.08 & 90.82 & 2.37 \\ 
         & $Dir=1.0$ & 314.71 & 80.89 & 1.49 \\
    \hline
        \multirow{4}{*}{CIFAR-100} & $Dir=0.1$ & 522.58 & 43.16 &  2.92 \\
         & $Dir=0.3$ & 238.08 & 38.07 &  1.32 \\
         & $Dir=0.5$ & 114.92 & 38.69 &  1.14 \\
         & $Dir=1.0$ & 72.59 & 42.77 &  1.10 \\
    \bottomrule
    \end{tabular}
\end{table}

\begin{figure}[t!]
    \centering
    \includegraphics[width=0.985\linewidth]{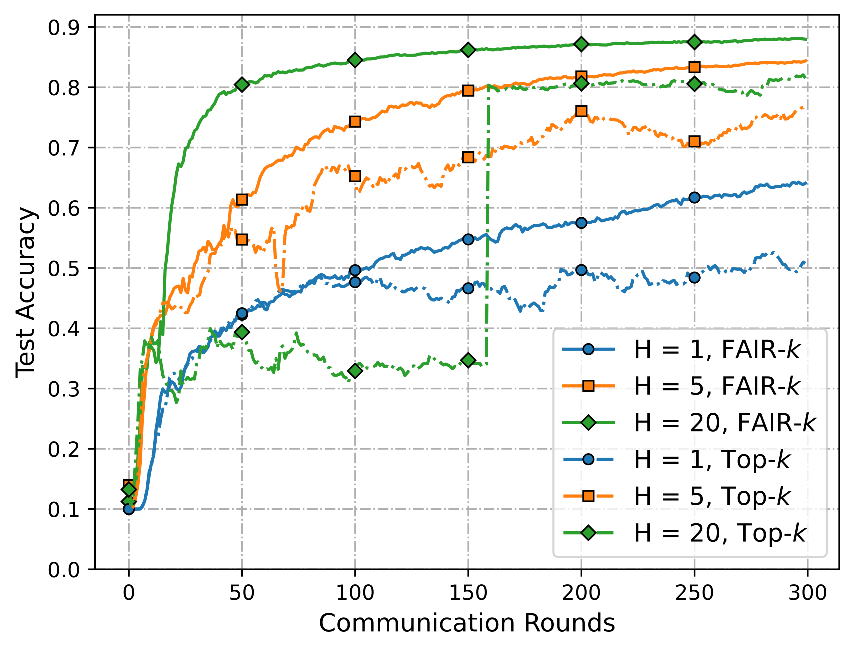}
    \caption{The impact of the local epoch, exemplified by training ResNet-18 on the CIFAR-10 dataset.}
    \label{fig:local epoch}
\end{figure}


\subsection{Prototype Demonstration}



\subsubsection{Settings}
To further demonstrate the efficacy of the proposed scheme, we implement FAIR-$k$ on a hardware prototype. 
The prototype comprises $N=2$ clients and one edge server, each implemented using an Adam-Pluto software-defined radio (SDR).
Each SDR is equipped with an AD9361 RF transceiver and a Xilinx Zynq XC7Z010 field-programmable gate array (FPGA) containing an embedded intellectual property (IP) core. 
A companion laptop handles baseband signal processing and model training, with independent software threads controlling each SDR.
An illustration of the prototype architecture is presented in Fig.~\ref{fig:prototype}. {\footnote{A video demonstration of our hardware prototype implementation is available at \url{https://www.youtube.com/watch?v=VcMUM2pbtA4}.}} 

\begin{figure}[t!]  
    \centering 
    \includegraphics[width=0.985\linewidth]{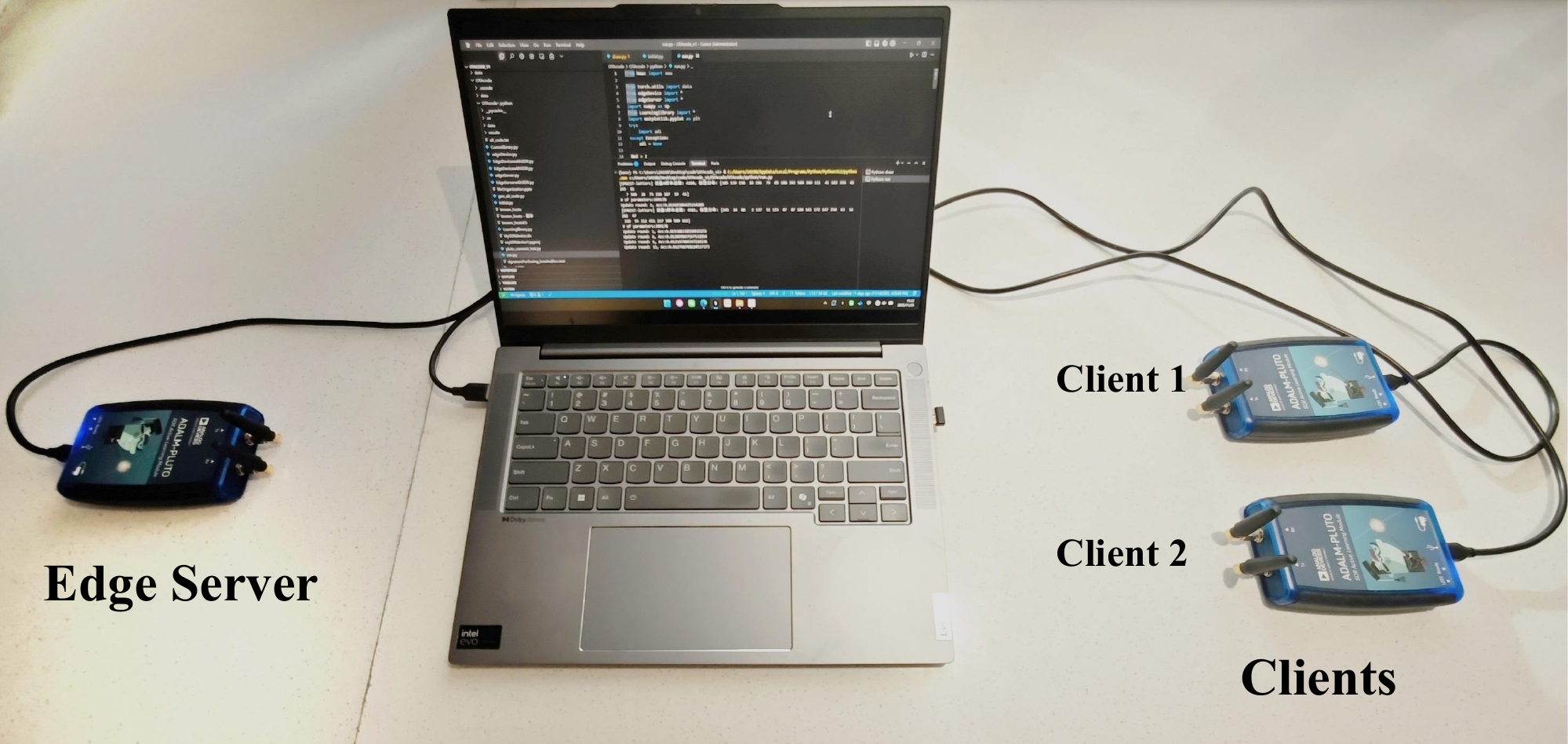}	
    \caption{System setup of the prototype.}
    \label{fig:prototype}
\end{figure}

\begin{figure}[t!]  
    \centering 
    \includegraphics[width=0.985\linewidth]{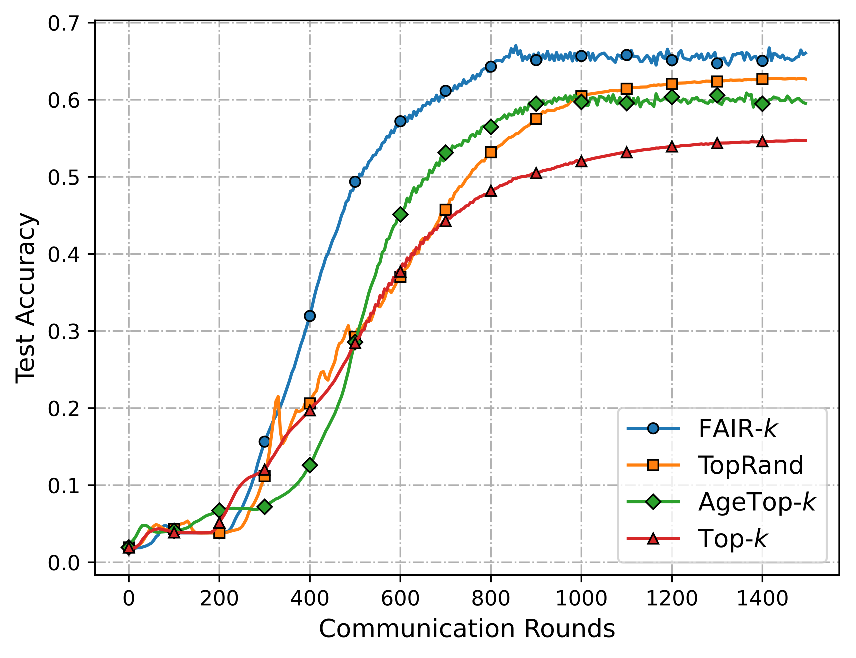}	
    \caption{Performance comparison based on prototype.}
    \label{fig:prototype_training}
\end{figure}

For hardware evaluation, we employ a CNN model \cite{lecun1998gradient} for image classification task on the EMNIST dataset \cite{cohen17emnist}.
The CNN model comprises three convolutional layers, followed by a fully connected layer with ReLU activation, with the total number of model parameters $d=109,402$.
The EMNIST letters dataset contains 145,600 data samples across 26 categories, with 124,800 for training and 20,800 for testing.
Due to hardware constraints, FAIR-$k$ is modified to incorporate one-bit quantization for gradient transmission.
More concretely, the local gradient of client $k$ is compressed as $\text{Sign}(\check{\boldsymbol{g}}_{n,t})$.
Upon receiving the aggregated compressed gradient $\check{\boldsymbol{g}}_{t}$, the edge server updates the sparse selection and applies the frequency-shift keying (FSK)-based majority vote (MV) \cite{csahin23distributed} strategy to determine the sign of $\check{\boldsymbol{g}}_{t}$, thereby quantifying each entry to one bit.
The reconstructed global gradient is then broadcast to all clients for model updates.

The experiments are conducted in a low-mobility indoor environment, where the clients and the edge server are separated by approximately $1.5$~m.
We employ $1,138$ OFDM symbols, organized into $4$ sequential groups, with $192$ active subcarriers utilized. 
The SDRs operate with a sampling rate of $20$~Msps and an FPGA clock frequency of $100$ MHz.
To achieve precise synchronization, we adopt the methods in \cite{csahin22demonstration}, which incorporate a hard-coded synchronization block within each SDR and a closed-loop calibration mechanism to correct time offsets caused by FPGA clock drifts.

\subsubsection{Experiment Results}
In Fig.~\ref{fig:prototype_training}, we plot the test accuracy collected from our hardware prototype to demonstrate the efficacy of the proposed FAIR-$k$ algorithm under the compression ratio $\rho=20\%$.
The results confirm that the prototype achieves reliable performance utilizing the FSK-MV strategy and precise synchronization, validating the functionality of OAC-FL systems in a practical implementation.
Moreover, the performance comparison derived from the prototype exhibits similar convergence trends as observed in the simulations.
FAIR-$k$ maintains superior and stable performance compared to baselines, demonstrating its robustness even when coupled with one-bit quantization.

\section{Conclusion}
In this paper, we proposed FAIR-$k$, an age-aware parameter updating scheme for refreshing the global model in OAC-FL systems, addressing the waveform-dimensionality disparity issue caused by the mismatch between limited waveform resources and high-dimensional model parameters.
By introducing the metric of AoU, FAIR-$k$ incorporates freshness information into the algorithm design, delicately balancing the importance and timeliness in each entry of the global model, hence being able to identify the most impactful subset of gradients to be updated in each communication round. 
We analyzed the impact of FAIR-$k$ on parameter staleness by modeling the parameter selection dynamics through a Markov transition process and derived the AoU distribution.
We further established the convergence rate of OAC-FL with FAIR-$k$, capturing the joint effect of data heterogeneity, parameter staleness, and channel noise on the model training efficiency.
We conducted extensive simulations to verify the effectiveness of FAIR-$k$ as well as implemented the algorithm in an OAC-FL prototype system using SDRs.
The results showed that FAIR-$k$ consistently outperformed the baselines, effectively mitigating average parameter staleness throughout training, and broadening the set of entries that receive fair opportunities to participate in model updates.
This work highlights the potential of incorporating information freshness metrics into machine learning algorithms, an aspect that is often overlooked but fundamentally important.

\section{Appendix}


\subsection{Proof of Lemma 1}

At communication round $t$, if the parameter staleness of a typical entry is $\tau=l$, it implies that starting from round $t-l-1$ (without loss of generality, we consider $l<t$), the entry has not been selected for update for $l$ consecutive communication rounds and is finally scheduled at communication round $t$. Equivalently, this corresponds to the entry starting at state $i$ and reaching either state $1$ or state $k_A+1$ for the first time after $l+1$ steps. The probability of this event can be expressed as
 \begin{align}
     &\mathbb{P}(\text{Entry reaches state $1$ or $k_A+1$ in $l$ steps from state $i$}) \nonumber \\
     &= \left(\mathbf{P}^l_{(1, k_A+1)}\mathbf{P}\right)_{i,1} + \left(\mathbf{P}^l_{(1, k_A+1)}\mathbf{P}\right)_{i,k_A+1}.
 \end{align}Because the probability of a typical entry being in state $i$ is given by $\pi_i$, the proof is complete by invoking the law of total probability.


\subsection{Proof of Lemma 2}
For ease of exposition, we introduce an auxiliary vector $\bar{\boldsymbol{g}}_t$ with the $i$-th entry being
\begin{small}
\begin{align}
    \bar{\boldsymbol{g}}_{t,i} = \frac{1}{N} \sum^N_{n=1} h_{n, t-\tau_i} \nabla \widetilde{f}_{n, i} (\boldsymbol{w}_{t-\tau_i}).
\end{align}
\end{small}

The randomness of $\bar{\boldsymbol{g}}_t$ stems from three aspects: (i) the staleness associated with the parameters, (ii) the stochastic data sampling during local updates, and (iii) the communication noise.
We thereby deal with the randomness as follows:
\begin{small}
\begin{align}\label{eq: g_t}
    &\mathbb{E}\left[\|\bar{\boldsymbol{g}}_t\|^2\right]
   \overset{(a)}{=}\mathbb{E}_{\tau_i, i\in[d]}\!\!\left[ \sum^d_{i=1} \!\left(\!\!\frac{1}{N} \!\sum^N_{n=1}\! h_{n, t-\tau_i} \nabla \widetilde{f}_{n, i} (\boldsymbol{w}_{t-\tau_i}) \!\!\right)^{\!2}\right] 
    \nonumber \\
    &=\sum^{\mathcal{T}}_{l=0} q_l \mathbb{E} \!\left[ \left\| \frac{1}{N} \!\sum^N_{n=1} h_{n,t-l} \!\sum^{H-1}_{s=0}\! \nabla {f}_{n} (\boldsymbol{w}_{n,t-l}^{(s)};\theta_n^{(s)}) \right\|^2\right] 
    \nonumber \\
    &\overset{(b)}{\leq} 2 \sum^{\mathcal{T}}_{l=0} q_l \mathbb{E} \!\left[ \left\| \frac{1}{N} \!\sum^N_{n=1} h_{n,t-l} \!\sum^{H-1}_{s=0}\! \left(\widetilde{f}_{n,t-l}^{(s)} \!-\! \nabla f_{n,t-l}^{(s)} \right) \right\|^2\right]
    \nonumber \\
    &\quad + 2 \sum^{\mathcal{T}}_{l=0} q_l \mathbb{E} \!\left[ \left\| \frac{1}{N} \!\sum^N_{n=1} h_{n,t-l} \!\sum^{H-1}_{s=0}\! \nabla f_{n,t-l}^{(s)}  \right\|^2\right] 
    \nonumber \\
    &\overset{(c)}{\leq} 2(\mu_c^2 \!+\! \sigma_c^2) \!\sum^{\mathcal{T}}_{l=0} q_l \mathbb{E} \!\left[ \left\| \frac{1}{N} \!\sum^N_{n=1} \!\sum^{H-1}_{s=0}\! \nabla f_{n,t-l}^{(s)}  \right\|^2\right]\!\!+\!  \frac{2H \sigma_s^2 (\mu_c^2 \!+\! \sigma_c^2)}{N}
\end{align}
\end{small}where (a) takes the expectation with respect to $\tau_i$, $i\in [d]$, to address the staleness in the parameters; (b) follows by denoting $\nabla \widetilde{f}_{n,t}^{(s)} = \nabla {f}_{n} (\boldsymbol{w}_{n,t}^{(s)};\theta_n^{(s)})$ and $\nabla f_{n,t}^{(s)} = \nabla {f}_{n} (\boldsymbol{w}_{n,t}^{(s)})$; and (c) arises from the fact that $\{h_{n,t}\}^N_{n=1}$ are i.i.d. and invokes Assumption~3.

The proof completes by substituting \eqref{eq: g_t} into $\mathbb{E}[\|\boldsymbol{g}_t\|^2]  = \mathbb{E}\left[\|\bar{\boldsymbol{g}}_t\|^2\right] + \frac{d \sigma_z^2}{N^2}$.

\subsection{Proof of Theorem 1}

Leveraging the smoothness of $f(\boldsymbol{w})$, we have 
\begin{small}
\begin{align}\label{eq:L-smooth}
    &\mathbb{E}\left[f(\boldsymbol{w}_{t+1})\right]
    \!\leq\! \mathbb{E}\left[f(\boldsymbol{w}_{t})\right] \!-\! \eta \mathbb{E}[\left\langle \nabla f(\boldsymbol{w}_{t}), \boldsymbol{g}_t \right\rangle] \!+\! \frac{L_g \eta^2}{2} \mathbb{E}\!\left[\|\boldsymbol{g}_t\|^2\right]. 
\end{align}
\end{small}

In what follows, we bound the terms on the right-hand side separately.
Whereby taking the expectation on $\tau_i$, we bound the second term on the right-hand side of \eqref{eq:L-smooth} as follows:
\begin{small}
\begin{align}\label{eq: Q_1+Q_2}
    &-\eta \mathbb{E}_{\tau_i, i\in[d]}\left\langle \nabla f(\boldsymbol{w}_{t}), \bar{\boldsymbol{g}}_t \right\rangle
    \nonumber \\
    &= -\eta \mu_c \!\sum^{\mathcal{T}}_{l=0} \!q_l  \mathbb{E} \!\left \langle \!\nabla f (\boldsymbol{w}_{t}), \frac{1}{N}\!\sum^N_{n=1} \!\sum^{H-1}_{s=0}\! \nabla f_{n,t-l}^{(s)}  \!\right \rangle
    \nonumber \\
    &= \underbrace{-\eta \mu_c \!\sum^{\mathcal{T}}_{l=0} \!q_l  \mathbb{E} \!\left \langle \!\nabla f (\boldsymbol{w}_{t})\!-\!\nabla f (\boldsymbol{w}_{t-l}), \frac{1}{N}\!\sum^N_{n=1} \!\sum^{H-1}_{s=0}\! \nabla f_{n,t-l}^{(s)}  \!\right \rangle}_{Q_1}
    \nonumber \\
    &\quad \underbrace{-\eta \mu_c \!\sum^{\mathcal{T}}_{l=0} \!q_l  \mathbb{E} \!\left \langle \!\nabla f (\boldsymbol{w}_{t-l}), \frac{1}{N}\!\sum^N_{n=1} \!\sum^{H-1}_{s=0}\! \nabla f_{n,t-l}^{(s)}  \!\right \rangle}_{Q_2}.
\end{align}
\end{small}We then bound $Q_1$ and $Q_2$ separately.
By using the Cauchy-Schwartz and AM-GM inequalities, it yields
\begin{small}
\begin{align}\label{eq: Q_1}
    &Q_1 \!=\!- \eta \mu_c \! \sum^{\mathcal{T}}_{l=0}\! q_l \!\!\!\sum^{t-1}_{j=t-l} \!\!\mathbb{E}\!\left\langle\!\! \nabla f(\boldsymbol{w}_{j}) \!-\!\! \nabla f (\boldsymbol{w}_{j+1}),  \frac{1}{N}\!\sum^N_{n=1} \!\!\sum^{H-1}_{s=0}\!\! \nabla \widetilde{f}_{n,t-l}^{(s)} \!\!\right\rangle\
    \nonumber \\
    &\leq \eta \mu_c \! \sum^{\mathcal{T}}_{l=0} \!q_l \!\!\!\sum^{t-1}_{j=t-l} \!\! L_g \mathbb{E}\|\boldsymbol{w}_{j} \!-\! \boldsymbol{w}_{j+1} \| \cdot \mathbb{E}\left\|\frac{1}{N}\!\sum^N_{n=1} \!\sum^{H-1}_{s=0}\! \nabla \widetilde{f}_{n,t-l}^{(s)} \right\|
    \nonumber \\
    &\!\overset{(a)}{\leq}\! \mu_c L_g \eta^2 \sum^{\mathcal{T}}_{l=0} l  q_l \left(\frac{d \sigma_z^2}{2N^2} \!+\! G^2 H^2 \Big(\frac{1}{2} \!+\! \mu_c^2 \!+\! \sigma_c^2 \Big) \right)
    \nonumber \\
    &=\mu_c L_g \mathbb{E}[\tau] \eta^2  \left(\frac{d \sigma_z^2}{2N^2} \!+\!  G^2 H^2 \Big(\frac{1}{2} \!+\! \mu_c^2 \!+\! \sigma_c^2 \Big) \right)
\end{align}
\end{small}where (a) adopts Assumption~4.
Then, denoted by $\bar{\boldsymbol{w}}^{(s)}_{t} = \frac{1}{N}\sum^N_{n=1}\boldsymbol{w}^{(s)}_{n,t}$, and using Assumption~1 and 2, we expand $Q_2$ as follows:
\begin{small}
\begin{align}\label{eq: Q_2}
    &Q_2 
    =\frac{\eta \mu_c}{2H} \!\sum^{\mathcal{T}}_{l=0} \!q_l \!\left(\! \mathbb{E}\!\left[\left\|\frac{1}{N}\!\sum^N_{n=1} \!\sum^{H-1}_{s=0}\! \left(\nabla f_{n,t-l}^{(s)} \!-\!\nabla f (\boldsymbol{w}_{t-l}) \right) \right\|^2\right] \right.
    \nonumber \\
    &\qquad \left. - H^2 \mathbb{E}\left[ \|\nabla f (\boldsymbol{w}_{t-l})\|^2\right] \!-\! \mathbb{E}\!\left[\left\|\frac{1}{N}\!\sum^N_{n=1} \!\sum^{H-1}_{s=0}\! \nabla f_{n,t-l}^{(s)} \right\|^2\right] \right)
    \nonumber \\
    &\leq \frac{\eta \mu_c}{2H} \!\sum^{\mathcal{T}}_{l=0} \!q_l \!\left( \frac{2H L_h^2}{N}\!\sum^{H-1}_{s=0} \!\sum^N_{n=1} \!\mathbb{E}\!\left[\left\| \boldsymbol{w}^{(s)}_{n,t-l} \!-\! \bar{\boldsymbol{w}}^{(s)}_{t-l} \right\|^2 \right]  \right.
    \nonumber \\
    &\quad  + 2H L_g^2 \!\sum^{H-1}_{s=0} \!\mathbb{E}\left[\left\|\bar{\boldsymbol{w}}^{(s)}_{t-l} \!-\! \boldsymbol{w}_{t-l}  \right\|^2\right] \!-\!H^2 \mathbb{E}\left[ \|\nabla f (\boldsymbol{w}_{t-l})\|^2\right] 
    \nonumber \\
    &\qquad \qquad \qquad \qquad \qquad \quad\left. - \mathbb{E}\!\left[\left\|\frac{1}{N}\!\sum^N_{n=1} \!\sum^{H-1}_{s=0}\! \nabla f_{n,t-l}^{(s)} \right\|^2\right] \right).
\end{align}
\end{small}Substituting \eqref{eq: Q_1} and \eqref{eq: Q_2} to \eqref{eq:L-smooth} and invoking Lemma~2, with $\eta \leq \frac{\mu_c}{2H L_g (\mu_c^2 + \sigma_c^2)}$, we have the following:
\begin{small}
\begin{align}\label{eq: f}
    &\mathbb{E}\left[f(\boldsymbol{w}_{t+1})\right]
    \leq \mathbb{E}\left[f(\boldsymbol{w}_{t})\right] 
    \!+\! \eta \mu_c \!\sum^{\mathcal{T}}_{l=0} \!q_l \Bigg(\!\!-\! \frac{H}{2} \mathbb{E}\!\left[ \|\nabla f (\boldsymbol{w}_{t-l})\|^2\right]     
    \nonumber \\
    &\!+\! \frac{ L_h^2}{N}\!\sum^{H-1}_{s=0} \!\sum^N_{n=1} \!\mathbb{E}\!\left[\left\| \boldsymbol{w}^{(s)}_{n,t-l} \!-\! \bar{\boldsymbol{w}}^{(s)}_{t-l} \right\|^2 \right] 
    \!\!+\!  L_g^2 \!\sum^{H-1}_{s=0} \!\mathbb{E}\!\left[\left\|\bar{\boldsymbol{w}}^{(s)}_{t-l} \!-\! \boldsymbol{w}_{t-l}  \right\|^2\right] \!\!\Bigg) 
    \nonumber \\    
    &\!+ \mu_c L_g \eta^2 \mathbb{E}[\tau] \left(\frac{d \sigma_z^2}{2N^2} \!+\! G^2 H^2(\frac{1}{2} \!+\! \mu_c^2 \!+\! \sigma_c^2)\right)
    \!+\! \frac{dL_g\eta^2 \sigma_z^2}{2N^2}
    \nonumber \\
    &\!+ \frac{H L_g \eta^2 \sigma_s^2 (\mu_c^2 \!+\! \sigma_c^2)}{N}.
\end{align}
\end{small}Following the steps in \cite{wang24new}, with $\eta_l \leq \frac{1}{2 \sqrt{3} H L_g}$, we have 
\begin{small}
\begin{align}\label{eq: Q_3}
    &\mathbb{E}\!\left[ \|\bar{\boldsymbol{w}}^{(s)}_t \!-\! \boldsymbol{w}_t\|^2\right] 
    \!\leq\! \frac{30H \eta_l^2 L_h^2}{N} \sum^{H-1}_{s=0} \sum^{N}_{n=1}\mathbb{E}\!\left[\| \boldsymbol{w}_{n,t}^{(s)}\!-\!\bar{\boldsymbol{w}}_t^{(s)}\|^2 \right] 
    \nonumber \\
    &\qquad \quad + \frac{5(H\!-\!1)\eta_l^2 \sigma^2_s}{N} \!+\! 30H(H\!-\!1)\eta_l^2 \mathbb{E}\!\left[ \|\nabla f(\boldsymbol{w}_t)\|^2 \right]. 
\end{align}
\end{small}With $\eta_l \leq \frac{1}{\sqrt{6H (L_g^2 + L_h^2) }}$ \cite{wang24new}, we further obtain
\begin{small}
\begin{align}\label{eq: Q_4}
    \frac{1}{N}\!\!\sum^{H-1}_{s=0}\! \sum^{N}_{n=1}\! \mathbb{E}\!\left[ \|\boldsymbol{w}_{n,t}^{(s)}\!-\!\bar{\boldsymbol{w}}_t^{(s)}\|^2\right] 
    \!\!\leq\! 12\eta_l^2 \sigma_g^2(H\!-\!1)^3 \!+\!4\eta_l^2 \sigma_s^2(H\!-\!1)^2.
\end{align}
\end{small}By setting the local stepsize as $\eta_l \leq \frac{1}{2\sqrt{30} H L_g}$, we substitute \eqref{eq: Q_3} and \eqref{eq: Q_4} into \eqref{eq: f} and arrive at the following:
\begin{small}
\begin{align}
    &\mathbb{E}\left[f(\boldsymbol{w}_{t+1})\right]
    \leq \mathbb{E}\left[f(\boldsymbol{w}_{t})\right] 
    \!-\! \frac{\eta \mu_cH}{4} \!\sum^{\mathcal{T}}_{l=0} q_l \mathbb{E}\!\left[ \|\nabla f(\boldsymbol{w}_{t-l})\|^2 \right]
    \nonumber \\
    &+ 18 \eta \mu_c L_h^2 \eta_l^2 \sigma_g^2 (H\!-\!1)^3 \!+ 6 \eta \mu_c L_h^2 \eta_l^2 \sigma_s^2 (H\!-\!1)^2    
    \nonumber \\
    &+\! \frac{5H(H\!-\!1) \eta \mu_c L_g^2 \sigma_s^2 \eta_l^2}{N}     
    \!+\! \frac{H L_g \eta^2 \sigma_s^2 (\mu_c^2 \!+\! \sigma_c^2)}{N} 
    \!+\! \frac{dL_g\eta^2 \sigma_z^2}{2N^2}
    \nonumber \\
    &+\!\mu_c L_g \mathbb{E}[\tau] \eta^2 \!\left(\!\frac{d \sigma_z^2}{2N^2} \!+\! G^2 H^2(\frac{1}{2} \!+\! \mu_c^2 \!+\! \sigma_c^2)\!\right)\!.
\end{align}
\end{small}Rearranging and telescoping the terms on both sides of the above, we obtain
\begin{small}
\begin{align}
    &\frac{1}{T} \sum^{T-1}_{t=0}\sum^{\mathcal{T}}_{l=0} q_l \mathbb{E}\!\left[ \|\nabla f(\boldsymbol{w}_{t-l})\|^2 \right]
    \leq \frac{4\big(f(\boldsymbol{w}_0) \!-\! f(\boldsymbol{w}^*)\big)}{\eta \mu_c H T}    
    \nonumber \\
    &+ 72L_h^2 \eta_l^2 \sigma_g^2 (H\!-\!1)^2 \!+ 4(H\!-\!1)\eta_l^2 \sigma_s^2 \!\left(\!6L_h^2 \!+\! \frac{5L_g^2}{N} \!\right)
    \!+\!\frac{2d L_g \eta \sigma_z^2}{\mu_c H N^2}
    \nonumber \\
    &+\! \frac{4 \eta L_g \sigma_s^2 (\mu_c^2 \!+\! \sigma_c^2 )}{\mu_c N}
    \!+\! \frac{4L_g \eta \mathbb{E}[\tau]}{H}\!\left(\!\frac{d \sigma_z^2}{2N^2} \!+\! G^2 H^2(\frac{1}{2} \!+\! \mu_c^2 \!+\! \sigma_c^2)\!\right).
\end{align}
\end{small}We complete the proof by evoking further algebraic manipulations.

\bibliographystyle{IEEEtran}
\bibliography{bib/StringDefinitions,bib/IEEEabrv,bib/Compression-OTA}

\end{document}